\documentclass{article}
\usepackage{graphics}
\usepackage{graphicx}

\input{epsf}        
\newcommand{\cepsffig}[1]{\begin{center}{\mbox{\epsffile{#1}}}\end{center}}

\begin{document}

\title{\textbf{Genetic Algorithms and Quantum Computation}}
\author{Gilson A. Giraldi, Renato Portugal, Ricardo N. Thess \thanks{%
National Laboratory for Scientific Computing, Petropolis, RJ, Brazil, 
\texttt{\{gilson,portugal,rnthess\}@lncc.br}}}
\date{}
\maketitle

\begin{abstract}
\noindent Recently, researchers have applied genetic algorithms (GAs) to
address some problems in quantum computation. Also, there has been some
works in the designing of genetic algorithms based on quantum theoretical
concepts and techniques. The so called Quantum Evolutionary Programming has
two major sub-areas: Quantum Inspired Genetic Algorithms (QIGAs) and Quantum
Genetic Algorithms (QGAs). The former adopts qubit chromosomes as
representations and employs quantum gates for the search of the best
solution. The later tries to solve a key question in this field: what GAs
will look like as an implementation on quantum hardware? As we shall see,
there is not a complete answer for this question. An important point for
QGAs is to build a quantum algorithm that takes advantage of both the GA and
quantum computing parallelism as well as true randomness provided by quantum
computers. In the first part of this paper we present a survey of the main
works in GAs plus quantum computing including also our works in this area.
Henceforth, we review some basic concepts in quantum computation and GAs and
emphasize their inherent parallelism. Next, we review the application of GAs
for learning quantum operators and circuit design. Then, quantum
evolutionary programming is considered. Finally, we present our current
research in this field and some perspectives. \\[2ex]
\noindent \textbf{Keywords}: \textit{Genetic Algorithms, Quantum Computing,
Evolutionary Strategies}.
\end{abstract}

\pagestyle{empty}

\pagestyle{empty} \thispagestyle{empty}

\section{Introduction}

Our aim in this paper is two-fold. Firstly, we review the main works in the
application of Genetic Algorithms (GAs) for quantum computing as well as in
the Quantum Evolutionary Programming. Secondly, based on this review, we
offer new perspectives in the area which are part of our current research in
this field.

In the last two decades we observed a growing interest in Quantum
Computation and Quantum Information due to the possibility to efficiently
solve hard problems for conventional computer science paradigms. Quantum
computation and quantum information encompass processing and transmission of
data stored in quantum states (see \cite{Oskin2002} and references therein).
In these fields, the computation is viewed as effected by the evolution of a
physical system, which is governed by unitary operators, according to the
Laws of Quantum Mechanics \cite{Chuang2000}. The basic unity information is
the \textit{qubit,} the counterpart in quantum computing to the classical $%
0-1$ bit. Quantum Computation and Quantum Information explore quantum
effects, like quantum parallelism, superposition of states and entanglement
in order to achieve a computational theory more efficient than the classical
ones. This has been demonstrated through quantum factoring and Grover's
algorithm for database search \cite{Hughes2001}.

On the other hand, Genetic Algorithms (\textbf{GAs}) is a rapidly expanding
area of current research. They were invented by John Holland in the 1960s 
\cite{Holland1975}. Simply stated, GAs are stochastic search algorithms
based on the mechanics of natural selection and natural genetics \cite
{Goldberg1989,Mitchell1996,Koza1992}. They have attracted people from a wide
variety of disciplines, mainly due to its capabilities for searching large
and non-linear spaces where traditional methods are not efficient \cite
{Goldberg1989}.

From this scenario, emerge the application of genetic algorithms for quantum
computation as well as evolutionary programming based on quantum theoretical
concepts and techniques. Despite of the fact that there are few works in
these subjects yet, it is an exciting area of research in the field of
evolutionary computation.

When applying GAs, people are attracted by their capabilities for searching
a solution in non-usual spaces. That is way people investigate the
application of GAs for learning quantum operators \cite
{ThessJCIS2003,Thess2003} and in the designing of quantum circuits \cite
{yabuki00genetic,Williams-Gray1999,rubinstein00evolving}.

Those works rely on the fundamental result for quantum computing that all
the computation can be expanded in a \textit{circuit} which nodes are the
universal gates \cite{Chuang2000}. These gates offer an expansion of an
unitary operator $U$ that evolves the system in order to perform some
computation \cite{Chuang2000,Hughes2001}. Thus, we are naturally in the face
of two classes of problems: (1) Given a set of functional points $S=\left\{
\left( x,y\right) \right\} $ find the operator $U$ such that $y=U\cdot x$;
(2) Given a problem, find a quantum circuit that solves it. The former was
formulated in the context of GAs for learning algorithms \cite
{ThessJCIS2003,Thess2003} while the latter through evolutionary strategies 
\cite{yabuki00genetic,Williams-Gray1999,rubinstein00evolving}.

In \cite{ThessJCIS2003} we proposed a method based on genetic algorithms to
learn linear operators. The method was applied for learning quantum
(unitary) operators. It was demonstrated that it overcomes the limitations
of the work proposed by Dan Ventura \cite{Ventura2000}, which resembles
basic methods in neural networks.

For the second class of problems, we found three schemes outlined by Spector 
\cite{Spector1999}, based on the traditional tree-based genetic programming 
\cite{Koza1992}, stackless and stack-based linear genome. Moreover, in
another scheme proposed by Williams and Gray \cite{Williams-Gray1999}, an
unitary matrix for a known quantum circuit is used to find possible
alternative circuits.

More close to our actual research are the works by Rubinstein \cite
{rubinstein00evolving} and Yabuki \cite{yabuki00genetic}. The production of
entangled states and the quantum teleportation were the target problems in
these works. In \cite{rubinstein00evolving} each gate is encoded through a
structure (gate structures) and a quantum circuit is represented as a list
of such structures. Then, genetic operators (crossover and mutation) are
applied in order to evolve a randomly chosen initial population of circuits.
A special fitness function was also proposed and the scheme applied to find
circuits for entangled states production. Yabuki and Iba follow a similar
philosophy in \cite{yabuki00genetic} but changing the circuit
representation. The quantum teleportation was the focused problem. In this
case, authors performed a circuit optimization; that is, they start with a 
\textit{seed circuit} that had eleven gates and obtained another one, with
just eight gates, but that performs the same computation. That is a variant
of the second class of problems.

By 1996, Narayan and Moore \cite{Narayan-Moore1998} introduced a novel
evolutionary computing method where concepts and principles of quantum
computing are used to inspire evolutionary strategies. It was the first
attempt towards quantum evolutionary programming. The basic approach is
inspired on the multiple universes view of quantum theory: each universe
contains its own population of chromosomes. The populations in each universe
obey identical classical rules and evolve in parallel. However, just after
classical crossover within each universe, the universes can interfere with
one another which induces some kind of crossover involving the chromosomes.

Narayan and Moore's method depends on non-standard interpretations of
Quantum Mechanics. The lack of a more formal analysis of the physical
concepts used brings difficulties to make the correlation between the
physics and the genetic algorithm itself. Consequently it does not offer
clues of the advantages of a quantum implementation for GAs within the
current implementations of quantum computers \cite{Oskin2002}. Thus, we are
not going to consider it in the following sections.

In \cite{han00genetic} we found another Quantum Inspired Genetic Algorithm
(QIGA) which relies on usual methods of Quantum Mechanics. It is
characterized by principles of quantum computing, including concepts of
qubits and superposition of states, as well as quantum operators to improve
convergence. An important consequence of this work is to emphasize that the
application of quantum computing concepts to evolutionary programming is a
promising field.

Rylander et al. \cite{rylander:2001:gecco} kept this idea and sketched out a
Quantum Genetic Algorithm (QGA) which takes advantage of both the quantum
computing and GAs parallelism. Despite of the lack of a mathematical
explanation about the physical realization of the algorithm, we will show
that its philosophy is close to an implementation in current experimental
architectures for quantum computers \cite{Oskin2002}. The key idea is to
explore the quantum effects of superposition and entanglement to create a
physical state that store individuals and their fitness. When measure the
fitness, the system collapses to a superposition of states that have that
observed fitness. Starting from this idea, we propose in section \ref
{Discussion} a QGA which can take advantage of both quantum computing and
GAs paradigms. We present its physical foundations and discuss its
advantages over classical GAs.

This paper is organized as follows. Firstly, we give the necessary
background in quantum computation (section \ref{QComp}) and genetic
algorithms (section \ref{GA}). Then, the review sections start. We begin
with the GAs applications in quantum computation. So, in section \ref{OurGA}
we describe our work on GAs for learning quantum operators. Section \ref
{Quantum-Design} presents the works on GAs for circuit design. Following, in
section \ref{QEC}, we offer the review of quantum evolutionary approaches.
The QIGA proposed in \cite{han00genetic} is presented on section \ref{QIGAs}%
. We end the review by presenting on section \ref{QGAs} the QGA proposed in 
\cite{rylander:2001:gecco}. In section \ref{Discussion} we discuss the
considered methods and describe some issues. In particular, we propose a
physical model for the QGA. Finally, we present the conclusions on section 
\ref{Concl}.

\section{Background in Quantum Computation and GAs \label{Background}}

\subsection{Quantum Computation \label{QComp}}

In practice, the most useful model for quantum computation is the Quantum
Circuit one \cite{Chuang2000,Preskill2001}. The basic information unit in
this model is a \textit{qubit } \cite{Chuang2000}, which can be considered a
superposition of two independent states $\mid 0\rangle $ and $\mid 1\rangle $
, denoted by $\mid \psi \rangle =\alpha _{0}\mid 0\rangle +\alpha _{1}\mid
1\rangle $, where $\alpha _{0},\alpha _{1}$ are complex numbers such that $%
\left| \alpha _{0}\right| ^{2}+\left| \alpha _{1}\right| ^{2}=1$. They are
interpreted as \textit{probability amplitudes} of the states $\mid 0\rangle $
and $\mid 1\rangle .$

A composed system with $n$ qubits is described using $N=2^{n}$ independent
states obtained through the tensor product of the Hilbert Space associated
with each qubit. Its physical realization is called a \textit{quantum
register}. The resulting space has a natural basis that can be denoted by:

\begin{equation}
\left\{ \mid i_{0}i_{1}...i_{n-1}\rangle ;\quad i_{j}\in \left\{ 0,1\right\}
,\quad j=0,1,...,n-1\right\} ,  \label{basis00}
\end{equation}
where we are using the \textit{Dirac notation} for vectors in Hilbert spaces.

This set can be indexed by $\mid i\rangle ;\quad i=0,1,...,N-1.$ Following
the Quantum Mechanics Postulates \cite{Chuang2000}, the state $\mid \psi
\rangle ,$ of a system, in any time $t,$ can be expanded as a \textit{%
superposition} of the basis states:

\begin{equation}
\mid \psi \rangle =\sum_{i=0}^{N-1}\alpha _{i}\mid i\rangle ;\quad
\sum_{i=0}^{N-1}\left| \alpha _{i}\right| ^{2}=1.  \label{superpos00}
\end{equation}

Entanglement is another important concept for quantum computation with no
classical counterpart. To understand it, a simple example is worthwhile.

Let us suppose that we have a composed system with two qubits. According to
the above explanation, the resulting Hilbert Space has $N=2^{2}$ independent
states.

Let the Hilbert Space associated with the first qubit (indexed by $1$)
denoted by $H_{1}$ and the Hilbert Space associated with the second qubit
(indexed by $2$) denoted by $H_{2}$. The computational basis for these
spaces are given by: $\left\{ \mid 0\rangle _{1},\mid 1\rangle _{1}\right\} $
and $\left\{ \mid 0\rangle _{2},\mid 1\rangle _{2}\right\} $, respectively.
If qubit 1 is in the state $\mid \psi \rangle _{1}=a_{10}\mid 0\rangle
_{1}+a_{11}\mid 1\rangle _{1}$ and qubit 2 in the state $\mid \psi \rangle
_{2}=a_{20}\mid 0\rangle _{2}+a_{21}\mid 1\rangle _{2}$, then the composed
system is in the state: $\mid \psi \rangle =\mid \psi \rangle _{1}\otimes
\mid \psi \rangle _{2}$, explicitly given by:

\begin{equation}
\mid \psi \rangle =\sum_{i,j\in \left\{ 0,1\right\} }a_{1i}a_{2j}\mid
i\rangle _{1}\otimes \mid j\rangle _{2}.  \label{entan00}
\end{equation}

Every state that can be represented by a tensor product $\mid \psi \rangle
_{1}\otimes \mid \psi \rangle _{2}$ belongs to the tensor product space $%
H_{1}\otimes H_{2}$. However, there are some states in $H_{1}\otimes H_{2}$
that can not be represented in the form $\mid \psi \rangle _{1}\otimes \mid
\psi \rangle _{2}$. They are called \textit{entangled states. }The Bell
state (or \textit{EPR pair}), denoted by $\mid \beta _{00}\rangle $, is a
very known example:

\begin{equation}
\mid \beta _{00}\rangle =\frac{1}{\sqrt{2}}\left( \mid 0\rangle _{1}\otimes
\mid 0\rangle _{2}+\mid 1\rangle _{1}\otimes \mid 1\rangle _{2}\right) .
\label{Bell00}
\end{equation}

Trying to represent this state as a tensor product $\mid \psi \rangle
_{1}\otimes \mid \psi \rangle _{2}$, with $\mid \psi \rangle _{1}\in H_{1}$
and $\mid \psi \rangle _{2}\in H_{2}$, produces an inconsistent linear
system without solution.

Entangled states are fundamental for teleportation \cite
{Furusawa1998,Chuang2000}. In recent years, there has been tremendous
efforts trying to better understand the properties of entanglement, not only
as a fundamental resource for the Nature, but also for quantum computation
and quantum information \cite{Bennett1993,brassard98teleportation}.

The computation unit in quantum circuits's model consists of quantum gates
which are unitary operators that evolve an initial state performing the
necessary \textit{computation}. A quantum computing algorithm can be
summarized in three steps: (1) Prepare the initial state; (2) A sequence of
(universal) quantum gates to evolve the system; (3) Quantum measurements.

From quantum mechanics theory, the last stage performs a \textit{collapse}
and only what we know in advance is the probability distribution associated
to the measurement operation. So, it is possible that the result obtained by
measuring the system should be post-processed to achieve the target (quantum
factoring (Chapter 6 of \cite{Preskill2001}) is a nice example).

More formally, the measurement in quantum mechanics is governed by the
following postulate \cite{Chuang2000}. Quantum Measurements are described by
a collection $\left\{ M_{m}\right\} $ of \textit{measurement operators}
satisfying ($\sum_{m}M_{m}^{\dagger }M_{m}=1$) acting on the state space of
the system being measured. If the state on the system is $\mid \psi \rangle
, $ given by expression (\ref{superpos00}), immediately before the
measurement then the probability that result $m$ occurs is given by:

\begin{equation}
p\left( m\right) =<\psi \mid M_{m}^{\dagger }M_{m}\mid \psi \rangle ,
\label{measure000}
\end{equation}
and the state of the system just after the measurement is:

\begin{equation}
\mid \psi \rangle _{after}=\frac{M_{m}\mid \psi \rangle }{\sqrt{<\psi \mid
M_{m}^{\dagger }M_{m}\mid \psi \rangle }}.  \label{measure001}
\end{equation}

The expression (\ref{measure001}) is the mathematical description of the
collapse due to the measure. A simple but important example is the
measurement of the state $\mid \psi \rangle ,$ given by expression (\ref
{superpos00}) in the computational basis. This is the measurement of the
system with $N$ outcomes defined by the operators $\left\{ \mid 0\rangle
<0\mid ,\mid 1\rangle <1\mid ,\mid 2\rangle <2\mid ,...,\mid N-1\rangle
<N-1\mid \right\} .$ Thus, applying equations (\ref{measure000})-(\ref
{measure001}) we find:

\begin{equation}
p\left( m\right) =\left| \alpha _{m}\right| ^{2},\quad \mid \psi \rangle
_{after}=\frac{\alpha _{m}}{\left| \alpha _{m}\right| }\mid m\rangle ,\quad
m=0,...,N-1.  \label{mesure002}
\end{equation}

Thus, the collapse becomes more evident.

Quantum parallelism is another fundamental feature of quantum computing. To
better explain it, let us take the Hadamard Operator ($H$) defined by:

\begin{equation}
H\mid 0\rangle =\frac{\mid 0\rangle +\mid 1\rangle }{\sqrt{2}},\quad H\mid
1\rangle =\frac{\mid 0\rangle -\mid 1\rangle }{\sqrt{2}}.
\label{parallel000}
\end{equation}

Now, we shall present another quantum operator which will be useful in the
following sections. Suppose $f:\left\{ 0,1\right\} \rightarrow \left\{
0,1\right\} $ a binary function with a one-bit domain. Now, define a quantum
operator $U_{f}:$ $H_{1}\otimes H_{2}\rightarrow H_{1}\otimes H_{2}$ such
that:

\begin{equation}
U_{f}\mid x\rangle \otimes \mid y\rangle =\mid x\rangle \otimes \mid y\oplus
f\left( x\right) \rangle ,  \label{parallel001}
\end{equation}
where the symbol $\oplus $ means addition modulo $2$. If we take the state $%
\mid \psi \rangle =\mid 0\rangle \otimes \mid 0\rangle $ and apply the
Hadamard operator over the second qubit ($I\otimes H$), followed by $U_{f}$
we obtain:

\begin{equation}
U_{f}\left( I\otimes H\right) \mid 0\rangle \otimes \mid 0\rangle =U_{f}\mid
0\rangle \otimes \left( \frac{\mid 0\rangle +\mid 1\rangle }{\sqrt{2}}%
\right) =\mid 0\rangle \otimes \left( \frac{\mid f\left( 0\right) \rangle
+\mid f\left( 1\right) \rangle }{\sqrt{2}}\right) .  \label{parallel003}
\end{equation}

This is a remarkable result because it contains information about $f\left(
0\right) $ and $f\left( 1\right) .$ It is almost as if we have evaluated $%
f\left( x\right) $ for two values of $x$ simultaneously! This feature is
known as quantum parallelism and can be generalized to functions on an
arbitrary number of bits.

To simplify notations, we will represent the tensor product $\mid i\rangle
\otimes \mid j\rangle $ by $\mid ij\rangle $ or $\mid i\rangle \mid j\rangle 
$, in what follows.

A fundamental result for quantum computing is that any unitary matrix $U$
which acts on a $d-$dimensional Hilbert space can be represented by a finite
set of unitary matrices (\textit{Universal Gates}) which act non-trivially
only on lower subspaces. The Hadamard operator defined in expression (\ref
{parallel000}) is an universal quantum gate. Another one is the $CNOT$
operator, defined as follows:

\begin{eqnarray}
CNOT &\mid &00\rangle =\mid 00\rangle ; \quad CNOT\mid 01\rangle =\mid
01\rangle ;  \label{cnot000} \\
CNOT &\mid &10\rangle =\mid 11\rangle ; \quad CNOT\mid 11\rangle =\mid
10\rangle ;  \nonumber
\end{eqnarray}
that is; the action of the operator is such that if the first qubit (\textit{%
control qubit}) is set to zero, then the second qubit (\textit{target}) is
left unchanged. Otherwise, the target qubit is flipped. Other universal
quantum gates can be found in \cite{Chuang2000}.

Given an operator in a Hilbert space, we can take its action over the
computational basis to get a matrix representation. For the Hadamard and
CNOT gates, we have the following representations:

\begin{equation}
CNOT=\left[ 
\begin{array}{llll}
1 & 0 & 0 & 0 \\ 
0 & 1 & 0 & 0 \\ 
0 & 0 & 0 & 1 \\ 
0 & 0 & 1 & 0
\end{array}
\right] ; \quad H=\frac{1}{\sqrt{2}}\left[ 
\begin{array}{ll}
1 & 1 \\ 
1 & -1
\end{array}
\right] ,  \label{cnot-hadamard}
\end{equation}
according to expressions (\ref{parallel000}) and (\ref{cnot000}),
respectively.

\subsection{Evolutionary Computation and GAs \label{GA}}

In the 1950s and the 1960s several computer scientists independently studied
evolutionary systems with the idea that evolution could be used as an
optimization tool for engineering problems. The idea in all these systems
was to evolve a population of candidate solutions for a given problem, using
operators inspired by natural genetic and natural selection.

Since then, three main areas evolved: evolution strategies, evolutionary
programming, and genetic algorithms. Nowadays, they form the backbone of the
field of evolutionary computation \cite{Mitchell1996,Adami1998}.

Genetic Algorithms (GAs) were invented by John Holland in the 1960s \cite
{Holland1975}. Holland's original goal was to formally study the phenomenon
of adaptation as it occurs in nature and to develop ways in which the
mechanism of natural adaptation might be imported into computer systems. In
Holland's work, GAs are presented as an abstraction of biological evolution
and a theoretical framework for adaptation under the GA is given. Holland's
GA is a method for moving from one population of \textit{chromosomes} to a
new one by using a kind of \textit{natural selection} together with the
genetic-inspired operators of crossover and mutation. Each chromosome
consists of \textit{genes }(bits in computer representation), each gene
being an instance of a particular \textit{allele} ($O$ or $1$).

Traditionally, these crossover and mutations are implemented as follows \cite
{Holland1975,Mitchell1996}.

\textit{Crossover: }Two parent chromosomes are taken to produce two child
chromosomes. Both parent chromosomes are split into left and a right
subchromosomes. The split position (\textit{crossover point}) is the same
for both parents. Then each child gets the left subchromosome of one parent
and the right subchromosome of the other parent. For example, if the parent
chromosomes are 011 10010 and 100 11110 and the crossover point is between
bits 3 and 4 (where bits are numbered from left to right starting at 1),
then the children are 01111110 and 100 10010.

\textit{Mutation: }: When a chromosome is taken for mutation, some genes are
randomly chosen to be modified. The corresponding bits are \textit{flipped}
from $0$ to $1$ or from $1$ to $0$.

These operations reveal the fact that GAs are inherently parallel
algorithms. GAs work by discovering the most adapted chromosomes,
emphasizing, and recombining their good ''building blocks'' through
operations that can be easily performed in parallel. This has been explored
in many works over the GA literature \cite
{belding95distributed,Parallel-han00genetic}.

Next we present a generalization of the $0-1$ case, in which the alleles are
real parameters \cite{wright91genetic}. It belongs to the class of
real-coded Genetic Optimization Algorithms and will be used in section \ref
{OurGA}.

Genetic Optimization Algorithms are stochastic search algorithms which are
used to search large, non-linear spaces where expert knowledge is lacking or
difficult to encode and where traditional optimization techniques fall short 
\cite{Goldberg1989}.

To design a standard genetic optimization algorithm, the following elements
are needed:

(1) A method for choosing the initial population;

(2) A ''scaling'' function that converts the objective function into a
non-negative fitness function;

(3) A selection function that computes the ''target sampling rate'' for each
individual. The target sampling rate of an individual is the desired
expected number of children for that individual.

(4) A sampling algorithm that uses the target sampling rates to choose which
individuals are allowed to reproduce.

(5) Reproduction operators that produce new individuals from old ones.

(6) A method for choosing the sequence in which reproduction operators will
be applied

For instance, in \cite{wright91genetic} each population member is
represented by a chromosome which is the parameter vector $x=\left(
x_{1},x_{2},...,x_{m}\right) \in \Re ^{m}$, each component $x_{i}$ being a
gene. Consequently, alleles are allowed to be real parameters. Thus, some
care should be taken to define these operators.

Mutations can be implemented as a perturbation of the chromosome. In \cite
{wright91genetic}, the authors chosen to make mutations only in coordinate
directions instead of to make in $\Re ^{m}$ due to the difficult to perform
global mutations compatible with the schemata theorem (it is a fundamental
result for GAs \cite{Holland1975,Goldberg1989}).

\begin{figure}[tbph]
\epsfxsize=7.0cm
\par
\begin{center}
{\mbox{\epsffile{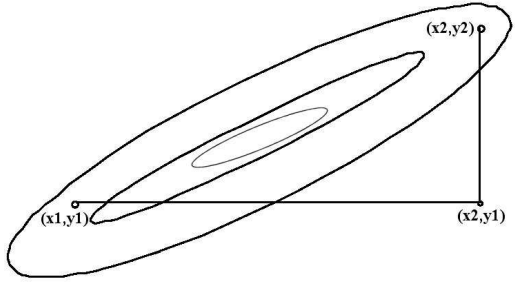}}}
\end{center}
\caption{Crossover can generate points out of the attraction region.}
\label{fig2}
\end{figure}

Besides, the crossover in $\Re ^{m}$ may also have problems. Figure \ref
{fig2} illustrates the difficult. The ellipses in the figure represent
contour lines of an objective function. A local minimum is at the center of
the inner ellipse. Points $(x_{1},y_{1})$ and $(x_{2},y_{2})$ are both
relatively good points in that their function value is not too much above
the local minimum. However, if we implement a traditional-like crossover
(section \ref{GA}) we may get points that are worse than their parents.

To address this problem, in \cite{wright91genetic} was proposed another form
of reproduction operator that was called linear crossover. From the two
parent points $p_{1}$ and $p_{2}$ three new points are generated, namely:

\[
\frac{1}{2}\left( p_{1}+p_{2}\right) ;\qquad \frac{3}{2}p_{1}-\frac{1}{2}%
p_{2};\qquad -\frac{1}{2}p_{1}+\frac{3}{2}p_{2}.\quad 
\]

The best two of these three points are selected.

Inspired on the above analysis we propose the algorithm of section \ref
{OurGA} to learn a linear operator from a set $S$ of example functional
points.

\section{Applying GAs for Quantum Computing \label{ApplyGA}}

\subsection{GA for Learning Operators \label{OurGA}}

\begin{sloppypar}Let us suppose that we do not know an operator $F:V\rightarrow V$ but, 
instead, we have a set
of functional points $S=\left\{ \left( \mid \chi _{i}\rangle ,\mid \psi
_{i}\rangle \right) ;\quad F\mid \chi _{i}\rangle =\mid \psi _{i}\rangle
,\quad i=0,1,...,n-1\right\} ,$ where $\dim \left( V\right) =n$, also 
called
the \textit{learning sequence}. We can hypothesize a function $G$ such that $%
\left\| G\mid \chi _{i}\rangle -\mid \psi _{i}\rangle \right\| \cong 0$ (as
usual, $\left\| \mid v\rangle \right\| =\sqrt{\langle v\mid v\rangle }$ is
the norm induced by the inner product).\end{sloppypar}

In this section, we present our general learning algorithm, based on GAs, to
find $G$ \cite{ThessJCIS2003}. This work was motivated by Dan Ventura's
algorithm for learning quantum operators \cite{Ventura2000}, which resembles
basic methods in neural networks \cite{Beale1994}. Our GA method has a range
of applications larger than that one of Dan Ventura's learning algorithm.
This is the main contribution of the work described next.

Following \cite{wright91genetic}, each population member (chromosome) is a
matrix $A\in \Re ^{n\times n}$, and alleles are real parameters (matrix
entries). The alleles are restricted to $\left[ -1,1\right] $, but more
general situations can be implemented.

The initial population is randomly generated. Once a population is obtained,
a fitness value is calculated for each member. The fitness function is
defined by: 
\begin{equation}
fitness\left( A\right) =\exp \left( -error\left( A\right) \right) ;\quad
A\in \Re ^{n\times n},  \label{genetic01}
\end{equation}
\begin{sloppypar}\noindent where the error function is defined as follows. Let the learning sequence 
$S=\left\{ \left( \mid \chi _{i}\rangle ,\mid \psi _{i}\rangle \right) ;\quad
i=1,...,m\right\} $ , then: \end{sloppypar} 
\begin{equation}
error\left( A\right) =\frac{1}{n\cdot m}\sum_{i=1}^{m}\left\| A\mid \chi
_{i}\rangle -\mid \psi _{i}\rangle \right\| _{1},  \label{genetic02}
\end{equation}
where $\left\| x\right\| _{1}$ denotes the 1-norm of a $x=\left(
x_{1},...,x_{n}\right) $, defined by: $\left\| x\right\| _{1}=\left|
x_{1}\right| +...+\left| x_{n}\right| .$

Once the fitness is calculated for each member, the population is sorted
into ascending order of the fitness values. Then, the GA loop starts. Before
enter the loop description, some parameters must be specified.

\textbf{Population Size}: Number of individuals in each generation ($N$).

\textbf{Elitism}: It might be convenient just to retain some number of the
best individuals of each population (members with best fitness) ($Ne$). The
other ones will be generated through mutation and/or crossover. This kind of
selection method was first introduced by Kenneth De Jong \cite{Mitchell1996}
and can improve the GA performance.

\textbf{Selection Pressure}: The degree to which highly fit individuals are
allowed many offsprings \cite{Mitchell1996} ($Ps$). For instance, for a
selection pressure of $0.6$ and a population with size $N$, we will get only
the $0.6*N$ best chromosomes to apply genetic operators.

\textbf{Mutation Number}: Maximum number of alleles that can undergo
mutation ($Nm$). We do not choose to make mutations (implemented as
perturbations) in $\Re ^{n\times n}$, likewise in \cite{wright91genetic}.
Instead, we randomly choose some matrix entries to be perturbed.

\textbf{Termination Condition}: Maximum number of generations ($Ngen$).

\textbf{Mutation and Crossover Probabilities}: $Pm$ and $Pc$, respectively. 
\newline

The crossover is defined as follows. Given two parents $A=\left[
a_{ij}\right] $ and $B=\left[ b_{ij}\right] $ \ the following steps are
performed until two offspring $C_{1},C_{2}$ are generated: (1) Randomly
choose one of the parents; (2) Take a $(ij)$ matrix entry and puts its value
on $c_{ij}$. Go to step (1).

The mutation is implemented as a perturbation of the alleles. Thus, given a
member $A$, the mutation operator works as follows: $A\rightarrow A+\Delta $%
; where $\Delta $ is a perturbation matrix. The mutation number establishes
the quantity of non-null entries for $\Delta .$ They are defined according
to the mutation probability and a pre-defined \textbf{Perturbation Size},
that is, a range $\left[ a,b\right] \in \Re $, such that $a\leq \Delta
_{ij}\leq b$.

Once the above parameters are pre-defined and the input set ($S$) is given,
the GA algorithm proceeds. In the following pseudo-code block, $P\left(
t\right) $ represents the population at the interaction time $t$ and $N$ is
its size. $Ngen$ is the maximum number of generations allowed, the procedure 
$Evaluate\_Sort$ calculates the fitness of each individual and sort the
chromosomes into ascending order of the fitness values. The integer $Ne$
defines de elite members, the parameter $Ps\in \left[ 0,1\right] $ defines
the selection pressure and $Nm$ the number of matrix entries that may
undergo mutations. \newline
\newline
\noindent Procedure Learning-GA \newline
$t\leftarrow 0$; \newline
initialize $P\left( t\right) $; \newline
while($t<Ngen$) do

$t\leftarrow t+1$;

$Evaluate\_Sort\left( P\left( t-1\right) \right) $;

Store in $P\left( t\right) $ the $Ne$ best members of $P\left( t-1\right) $;

Complete $P\left( t\right) $ by crossover and mutation; \newline
end while

\subsubsection{Experimental Results \label{Experimental}}

Firstly, we analyze the behavior of the GA learning algorithm for the same
example presented in \cite{Ventura2000}. The set $S$ is given by:

\begin{equation}
S=\left\{ 
\begin{array}{l}
\left( \frac{1}{\sqrt{5}}\left[ 
\begin{array}{l}
2 \\ 
1
\end{array}
\right] ,\frac{1}{\sqrt{10}}\left[ 
\begin{array}{l}
3 \\ 
1
\end{array}
\right] \right) \\ 
\left( \frac{1}{\sqrt{20}}\left[ 
\begin{array}{l}
-2 \\ 
4
\end{array}
\right] ,\frac{1}{\sqrt{40}}\left[ 
\begin{array}{l}
2 \\ 
-6
\end{array}
\right] \right)
\end{array}
\right\} ,  \label{quant02n}
\end{equation}
and the target is the Hadamard Transform $H$, defined by expression (\ref
{parallel000}).

The GA result over $25$ runs was always the correct one. The set of
parameters is given in the first line of Table \ref{Table1}. The
perturbation size is given by $\left[ 0.001,0.1\right] $.

\begin{table}[!htb]
\begin{center}
\begin{tabular}{lllllllll}
Matrix & $Ngen$ & $N$ & $Pc$ & $Pm$ & $Ps$ & $Ne$ & $Nm$ &  \\ 
$2\times 2$ & $100$ & $200$ & $0.85$ & $0.95$ & $0.30$ & $30$ & $1$ &  \\ 
$2\times 2$ & $200$ & $200$ & $0.85$ & $0.95$ & $0.30$ & $30$ & $1$ & 
\end{tabular}
\end{center}
\caption{Parameters for the examples of this section (definitions on section 
\ref{OurGA}).}
\label{Table1}
\end{table}

\begin{figure}[tbph]
\epsfxsize=7.0cm
\par
\begin{center}
{\mbox{\epsffile{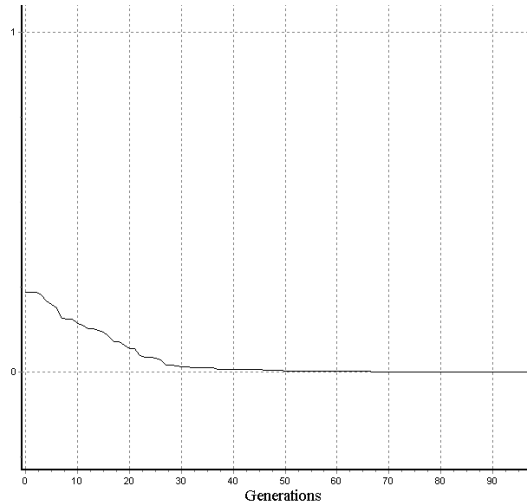}}}
\end{center}
\caption{Error evolution, over $25$ runs, for example given by expression (%
\ref{quant02n}).}
\label{caso2x2Ventura}
\end{figure}

Figure \ref{caso2x2Ventura} shows the error evolution over the $25$ runs. We
collect the best population member (smallest error) for each run and take
the mean value, for each generation, over the $25$ runs. It suggests that
the algorithm gets closer the solution fast but takes much more time to
achieve the target. Indeed, this behavior was observed for all experiments
reported in \cite{Thess2003}.

Dan Ventura's algorithm gives also the correct result for this example, as
reported in \cite{Ventura2000}.

Now, let us take the following set $S$:

\begin{equation}
S=\left\{ 
\begin{array}{l}
\left( \frac{1}{\sqrt{5}}\left[ 
\begin{array}{l}
1 \\ 
2
\end{array}
\right] ,\frac{1}{\sqrt{10}}\left[ 
\begin{array}{l}
3 \\ 
-1
\end{array}
\right] \right) \\ 
\left( \frac{1}{\sqrt{2}}\left[ 
\begin{array}{l}
1 \\ 
1
\end{array}
\right] ,\frac{1}{\sqrt{4}}\left[ 
\begin{array}{l}
2 \\ 
0
\end{array}
\right] \right)
\end{array}
\right\} .  \label{example002}
\end{equation}

In this set, the input vectors are not orthonormal ones. Thus, as we
demonstrate in \cite{ThessJCIS2003}, if we apply Ventura's algorithm we get
a result which is far from the target (see \cite{Thess2003} for more
details). But, our GA algorithm was able to deal with this case. The
operator to be learned is the Hadamard Transform, as before.

The second line of Table \ref{Table1} shows the parameters used. We would
like to keep all parameters unchanged but the number of generations ($Ngen$)
had to be increased to achieve the correct result. This point out that our
GA method may be sensitive to the fact that the set $\left\{ \mid \chi
_{0}\rangle ,\mid \chi _{1}\rangle \right\} $ is not an orthonormal basis,
despite that it learns correctly. Additional examples must be performed in
order to verify this observation.

The mean error evolution shows a behavior which is similar to the first
example. It decays fast but takes some time to become null.

Table \ref{Table1} shows that we could keep the crossover and mutation
probabilities unchanged for both these experiments (more results are
presented on \cite{Thess2003}). It is desired because it may indicates some
parameter stability. Moreover, the clock time for one run is very acceptable
($\leq 0.04$ seconds).

\subsection{GA for Quantum Circuit Design \label{Quantum-Design}}

Despite of the scientific and technological importance of quantum
computation, few quantum algorithms faster then the classical ones have been
discovered. Shor's algorithm for quantum factoring, Grove's quantum search
and Deutsch-Jozsa algorithm are basically the known ones \cite
{Hughes2001,Chuang2000}.

This is due to the fact that the generation of such algorithms or circuits
is difficult for a human researcher. They are unintuitive, mainly due to
quantum mechanics features like entanglement and collapse (section \ref
{QComp}). That is way researchers have investigated the use of stochastic
search techniques, such as genetic programming and genetic algorithms to
help in this task.

Among the main works in this subject \cite
{yabuki00genetic,rubinstein00evolving,Williams-Gray1999,Spector1999}, those
ones proposed in \cite{yabuki00genetic} and \cite{rubinstein00evolving} are
closer to our current research in this field.

\begin{figure}[htbp]
\epsfxsize=11.0cm
\par
\begin{center}
\begin{minipage}[b]{5.5cm}
    \begin{center}
      \epsfxsize=5.5cm
      \cepsffig{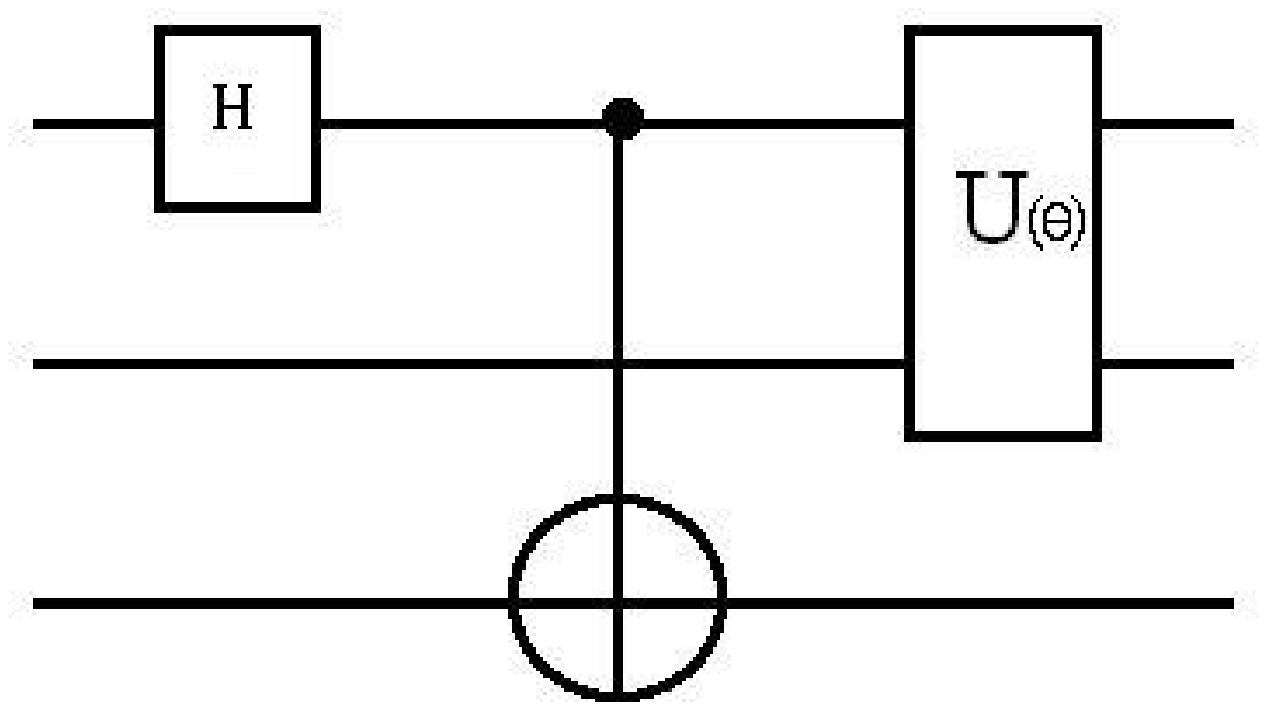}
    (a)
    \end{center}
  \end{minipage}
\begin{minipage}[b]{5.5cm}
    \begin{center}
      \epsfxsize=5.5cm
      \cepsffig{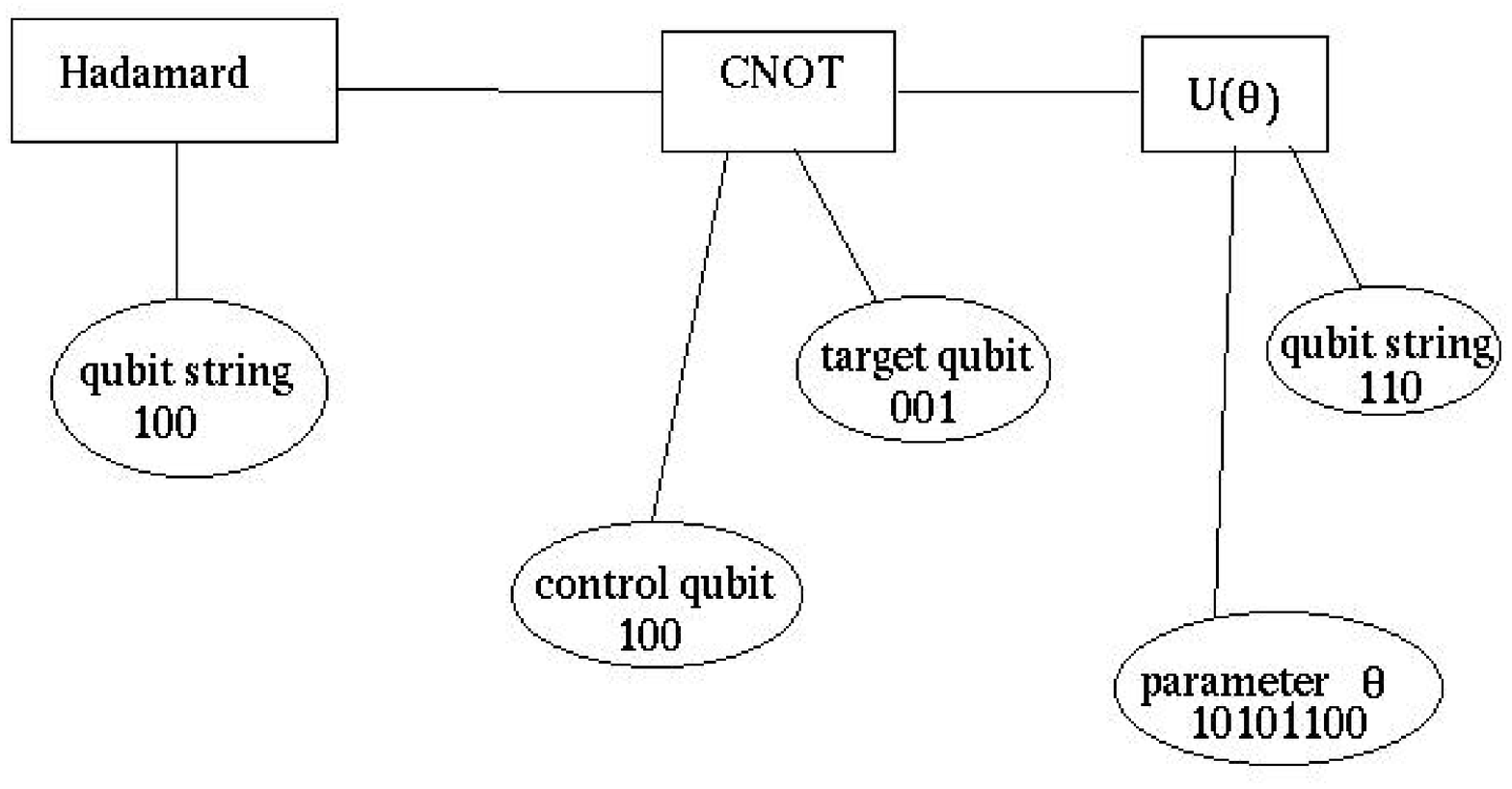}
    (b)
    \end{center}
  \end{minipage}
\end{center}
\caption{(a) Quantum Circuit. From the left to right:Hadamard ($H$), CNOT
and the Rotation Gate $U\left( \theta \right) $. (b) The corresponding
representation.}
\label{fig2-evolving-qc}
\end{figure}

The paper \cite{rubinstein00evolving} presents a new representation and
corresponding set of genetic operators for a scheme to evolve quantum
circuits. A quantum circuit is represented as a list of structures (gate
structures), where the size of a circuit (number of gates) can vary up to a
pre-defined maximum number. Each gate structure contains the gate type,
which is one in an allowable set of gate types including the usual Identity,
CNOT, Hadamard and measurement operators, and a binary string for each
category of qubit and parameter for that gate (see Figure{\ }\ref
{fig2-evolving-qc}).

The crossover and mutation operators are defined as follows. Crossover
operates on all the levels of an individual's structure: the gates, qubit
operands and parameter type. Gate crossover between two parent circuits
consists of picking a gate from each parent at random, and then swapping all
gates between the parents after these two points. Crossover between binary
strings representing parameters can only occur between strings of like
category, and proceeds in the same way as the crossover operator for the
fixed length GA: pick a crossing point, and then swap bit values between the
two strings after the point.

Mutation happens in the gate level. A gate is mutated by replacing it with a
new one randomly selected. Mutation is performed with small probability
(typically 0.001) because in \cite{rubinstein00evolving} mutation is
considered more an insurance against loss of important building blocks than
a fundamental search procedure. Such viewpoint may be changed for circuit
optimization. We shall return to this point ahead.

An error function is defined to compare the stored state(s) with the desired
one(s): Given a set of cases consisting of input states and desired outputs,
the error is defined by $error=\sum_{i,j}\left| \sigma _{ij}-d_{ij}\right| $%
, where we take for each case $i$, the sum of the magnitudes of the
differences between the probability amplitudes of the desired result $d_{i}$
and that obtained one $\sigma _{i}$. A fitness function is constructed based
on this error \cite{rubinstein00evolving}. The production of entangled
states was the focused application.

Reference \cite{yabuki00genetic} is another proposal in the application of
GAs for quantum circuit design. The philosophy is similar to the one
presented above.

The case study is the teleportation circuit \cite{brassard98teleportation}.
Quantum teleportation is a technique by which a quantum state can be
transported from one point to another through non-local interactions \cite
{Bennett1993}. To illustrate the steps involved in quantum teleportation let
us consider that there are two friends, Alice and Bob.

Imagine that Alice wants to deliver a qubit $\mid f\rangle =p\mid 0\rangle
+q\mid 1\rangle $ \ to Bob, who lives far apart. Alice does not know $p$ and 
$q.$ Moreover, she can not have access to these values by measuring the
qubit because, according to the quantum mechanics postulates, the system
will collapse to a state $p/\left| p\right| \mid 0\rangle $ or $q/\left|
q\right| \mid 1\rangle $. \ Once measurement is an irreversible task, the
information would be lost.

\begin{figure}[tbph]
\epsfxsize=8.0cm
\par
\begin{center}
{\mbox{\epsffile{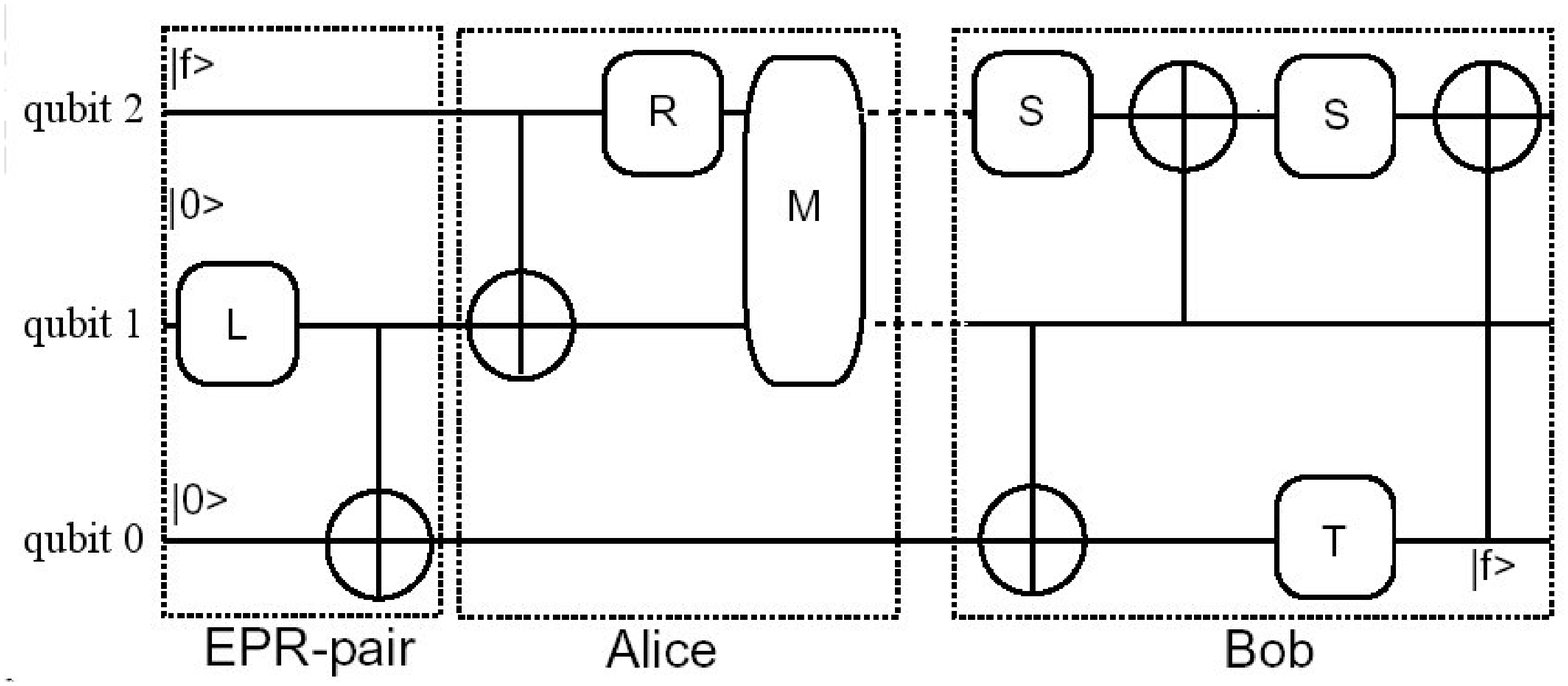}}}
\end{center}
\caption{Brassard's circuit for teleporting a single qubit.}
\label{fig2-design-qc}
\end{figure}

The scheme to solve this problem comprises the quantum teleportation. Its
physical basis was proposed by Bennet et al. \cite{Bennett1993}, followed by
Brassard \cite{brassard98teleportation}, who proposed the quantum circuit of
Figure \ref{fig2-design-qc} for teleporting a single qubit.

In the circuit of Figure \ref{fig2-design-qc}, we have the following
components: quantum gates $L,$ $R,$ $S,$ $T$, given by:

\begin{equation}
L=\frac{1}{\sqrt{2}}\left[ 
\begin{array}{cc}
1 & -1 \\ 
1 & 1
\end{array}
\right] , \quad R=\frac{1}{\sqrt{2}}\left[ 
\begin{array}{cc}
1 & 1 \\ 
-1 & 1
\end{array}
\right] ,  \label{teleport000}
\end{equation}

\begin{equation}
S=\left[ 
\begin{array}{cc}
i & 0 \\ 
0 & 1
\end{array}
\right] ,\quad T=\left[ 
\begin{array}{cc}
-1 & 0 \\ 
0 & -i
\end{array}
\right] ,  \label{teleport001}
\end{equation}
measurement operator $M$ and the CNOT gate, already defined in expression (%
\ref{cnot000}), and represented like in Figure \ref{fig2-evolving-qc}.

The circuit has $3$ qubits, namely qubit $0,1$ and $2$, from the bottom to
the top of the circuit. The method explores the concept of entanglement by
using the \textit{EPR }state given in equation (\ref{Bell00}). The first
part of the circuit is used to create that kind of entangled state. In this
operation, the zeroth and first qubits are affected. The input state is
given by the tensor product $\mid f\rangle \mid 0\rangle \mid 0\rangle $ and
the computation performs as follows:

\[
I\otimes L\otimes I\mid f\rangle \mid 0\rangle \mid 0\rangle =\mid f\rangle
\left( \frac{\mid 0\rangle +\mid 1\rangle }{\sqrt{2}}\right) \mid 0\rangle
=\mid f\rangle \left( \frac{\mid 0\rangle \mid 0\rangle +\mid 1\rangle \mid
0\rangle }{\sqrt{2}}\right) 
\]

Then, by applying the $CNOT_{01}$; that is, the CNOT gate, defined by
equations (\ref{cnot000}), with the first qubit as the control one, we find:

\begin{equation}
I\otimes CNOT_{01}\left[ \mid f\rangle \left( \frac{\mid 0\rangle \mid
0\rangle +\mid 1\rangle \mid 0\rangle }{\sqrt{2}}\right) \right] =\mid
f\rangle \left( \frac{\mid 0\rangle \mid 0\rangle +\mid 1\rangle \mid
1\rangle }{\sqrt{2}}\right) \equiv \mid f\rangle \mid \beta _{00}\rangle ,
\label{epr-criate000}
\end{equation}
where $\mid \beta _{00}\rangle $ is the Bell state defined in expression (%
\ref{Bell00}). Similarly, the Alice's circuit operates on the state given by
expression (\ref{epr-criate000}). The result has the general form:

\begin{equation}
\mid \psi \rangle _{After-Alice}=\sum_{i,j,k\in \{0,1\}}a_{i,j,k}\mid
ijk\rangle ,  \label{alice000}
\end{equation}

Then, Alice measures the first and second qubits. Thus, the state just after
the measurement will be one of the:

\begin{eqnarray}
&\mid &\psi \rangle _{0}=a_{000}\mid 000\rangle +a_{001}\mid 001\rangle ;
\quad \mid \psi \rangle _{2}=a_{100}\mid 100\rangle +a_{101}\mid 101\rangle ;
\label{alice001} \\
&\mid &\psi \rangle _{1}=a_{010}\mid 010\rangle +a_{011}\mid 011\rangle ;
\quad \mid \psi \rangle _{3}=a_{110}\mid 110\rangle +a_{111}\mid 111\rangle .
\nonumber
\end{eqnarray}

Each result will be processed by Bob's part. If we trace each measurement
result we can confirm that the initial state of the second qubit was
delivered to the zeroth qubit, which belongs to Bob. For instance, let us
suppose that the result was $\mid \psi \rangle _{0}$ given above. Then,
Bob's circuit will outputs the following state \cite{yabuki00genetic}:

\[
\left( \mid 0\rangle \mid 0\rangle -\mid 1\rangle \mid 0\rangle \right)
\left( p\mid 0\rangle +q\mid 1\rangle \right) \equiv \left( \mid 0\rangle
\mid 0\rangle -\mid 1\rangle \mid 0\rangle \right) \mid f\rangle , 
\]
which is a desired one, once the state of the second qubit was delivered to
the zeroth; that is, it was teleported. The final state of the second qubit
is different from the original one, which is accordance with the non-cloning
theorem of quantum mechanics \cite{Chuang2000}.

Once we have a quantum circuit (Figure \ref{fig2-design-qc}) that performs
the required computation, an interesting question arises: Given that quantum
circuit how to find another one which performs the same computation but has
less elementary gates? This \textit{optimization} problem was addressed in 
\cite{yabuki00genetic} through genetic algorithms (GAs).

In fact, in \cite{yabuki00genetic}, a circuit for quantum teleportation is
encoded by a chromosome that is a string of integers chosen from the set $%
\left\{ 0,1,2,3\right\} .$ Each gene is interpreted with a codon, i.e., a
three-letter unit. The first letter indicates a kind of gate, whereas the
second and the third letters indicate the qubits that the gate will operate.
For instance, let us consider the following string:

\begin{equation}
112\cdot 231\cdot 001\cdot \mathbf{3}31\cdot 132\cdot 012\cdot 122\cdot 
\mathbf{3}02\cdot 203\cdot 220\cdot 020\cdot 001  \label{string000}
\end{equation}

The first codon whose first letter is $3$ is interpreted as the partition
between EPR-pair generation and Alice's part. The second codon whose first
letter is $3$ corresponds to Alice's measurement. The codification used in 
\cite{yabuki00genetic} is defined by Tables \ref{Table-EPR-Alice} and \ref
{Table-Bob}:

\begin{table}[!htb]
\begin{center}
\begin{tabular}{llllll}
& 0 & 1 & 2 & 3 &  \\ 
0 & $CNOT_{01}$ & $CNOT_{10}$ &  &  & 0 \\ 
0 & $CNOT_{01}$ & $CNOT_{10}$ &  &  & 1 \\ 
0 & $CNOT_{01}$ & $CNOT_{10}$ &  &  & 2 \\ 
0 &  &  &  &  & 3 \\ 
1 & $L_{0}$ & $L_{1}$ &  &  & 0 \\ 
1 & $L_{0}$ & $L_{1}$ &  &  & 1 \\ 
1 & $L_{0}$ & $L_{1}$ &  &  & 2 \\ 
1 &  &  &  &  & 3 \\ 
2 & $R_{0}$ & $R_{1}$ &  &  & 0 \\ 
2 & $R_{0}$ & $R_{1}$ &  &  & 1 \\ 
2 & $R_{0}$ & $R_{1}$ &  &  & 2 \\ 
2 &  &  &  &  & 3 \\ 
3 & $\cdot $ & $\cdot $ & $\cdot $ & $\cdot $ & *
\end{tabular}
\begin{tabular}{llllll}
& 0 & 1 & 2 & 3 &  \\ 
0 & $CNOT_{12}$ & $CNOT_{21}$ &  &  & 0 \\ 
0 & $CNOT_{12}$ & $CNOT_{21}$ &  &  & 1 \\ 
0 & $CNOT_{12}$ & $CNOT_{21}$ &  &  & 2 \\ 
0 &  &  &  &  & 3 \\ 
1 & $L_{1}$ & $L_{2}$ &  &  & 0 \\ 
1 & $L_{1}$ & $L_{2}$ &  &  & 1 \\ 
1 & $L_{1}$ & $L_{2}$ &  &  & 2 \\ 
1 &  &  &  &  & 3 \\ 
2 & $R_{1}$ & $R_{2}$ &  &  & 0 \\ 
2 & $R_{1}$ & $R_{2}$ &  &  & 1 \\ 
2 & $R_{1}$ & $R_{2}$ &  &  & 2 \\ 
2 &  &  &  &  & 3 \\ 
3 & $\cdot $ & $\cdot $ & $\cdot $ & $\cdot $ & *
\end{tabular}
\end{center}
\caption{Codification for EPR generation (left). Alice's gates codification
(right).}
\label{Table-EPR-Alice}
\end{table}

\begin{table}[!htb]
\begin{center}
\begin{tabular}{llllll}
& 0 & 1 & 2 & 3 &  \\ 
0 &  & $CNOT_{10}$ & $CNOT_{20}$ &  & 0 \\ 
0 & $CNOT_{01}$ &  & $CNOT_{21}$ &  & 1 \\ 
0 & $CNOT_{02}$ & $CNOT_{12}$ &  &  & 2 \\ 
0 &  &  &  &  & 3 \\ 
1 & $L_{0}$ & $L_{1}$ & $L_{2}$ &  & 0 \\ 
1 & $L_{0}$ & $L_{1}$ & $L_{2}$ &  & 1 \\ 
1 & $L_{0}$ & $L_{1}$ & $L_{2}$ &  & 2 \\ 
1 &  &  &  &  & 3 \\ 
2 & $R_{0}$ & $R_{1}$ & $R_{2}$ &  & 0 \\ 
2 & $R_{0}$ & $R_{1}$ & $R_{2}$ &  & 1 \\ 
2 & $R_{0}$ & $R_{1}$ & $R_{2}$ &  & 2 \\ 
2 &  &  &  &  & 3 \\ 
3 & $\cdot $ & $\cdot $ & $\cdot $ & $\cdot $ & *
\end{tabular}
\end{center}
\caption{Codons for Bob's circuit generation.}
\label{Table-Bob}
\end{table}

This circuit representation is closer that one proposed in \cite
{rubinstein00evolving} (described above). However, differently from that
reference, in this representation the same integer may have different
meaning in different parts of the chromosome. For example, the letter $0$
means $CNOT_{01}$ during the EPR-pair generation while it means $CNOT_{12}$
for Alice's part.

Once established the circuit representation, genetic operators based on
mutations and crossover shall be specified. Mutations are implemented by
properly change the alleles of a given chromosome. A two-point crossover is
implemented by randomly choosing two parent chromosomes and exchange their
alleles.

Finally, the following steps are executed: (1) Decode each chromosome in a
circuit and its gates; (2) Apply these transformation on the initial state;
(3) If the circuit outputs a final state similar to the desired one, its
fitness is enlarged. (4) Apply the genetic operators to generate the next
population; (5) Go to step (1).

Each individual (circuit) is evaluated as follows:

(1) Make three random numbers $\alpha ,$ $\beta ,$ $\gamma \in \left[ 0,2\pi
\right] .$

\begin{sloppypar}(2) Prepare three initial states $\left( p,q\right) $ given by: $\left(
e^{i\beta }\cos \alpha ,e^{i\gamma }\sin \alpha \right) ,$ $\left(
e^{i\gamma }\cos \beta ,e^{i\alpha }\sin \beta \right) ,\left( e^{i\alpha
}\cos \gamma ,e^{i\beta }\sin \gamma \right) $ \end{sloppypar} 

(3) Derive the circuit by decoding the string using Tables \ref
{Table-EPR-Alice} and \ref{Table-Bob}.

(4) Use the state $\left( p\mid 0\rangle +q\mid 1\rangle \right) \otimes
\mid 00\rangle $ as the input one.

(5) Evaluate the fitness.

(6) Change random numbers every 50 generations.

After the measurement, we have to trace all the branches. corresponding to
the possible outcomes given by equations (\ref{alice001}). The desired final
state, at the end of Bob's circuit has a general form $\left( a\mid
00\rangle +b\mid 01\rangle +c\mid 10\rangle +d\mid 11\rangle \right) \otimes
\left( p\mid 0\rangle +q\mid 1\rangle \right) \equiv \left(
ap,aq,bp,bq,cp,cq,dp,dq\right) .$ Thus, the ratio $a_{i}/a_{i+1}=p/q;$ $%
i=0,2,4,6.$ This gives a clue to find out an efficient fitness function.

Given a circuit, we observe that each one of the initial states produces $4$
final states (one for each possible measurement's result, given by
expressions (\ref{alice001})). Thus, for the three initial states, we will
have $12$ possible final states. If we write the final states as $\left(
a_{j,0};a_{j,1};...,a_{j,7}\right) $, $j=0,2,..,11$, we can express the gap
between a final state and the desired one as:

\[
error_{j}=\frac{1}{n}\sum_{i=0,2,4,6}\left| \frac{a_{j,i}}{a_{j,i+1}}-\frac{p%
}{q}\right| , 
\]
where $n$ is the numbers of pairs such that $\left( a_{j,i},a_{j,i+1}\right)
\neq \left( 0,0\right) $ and the summation is taken over such pairs. In the
case that the final state is $\mathbf{0}$ then $error_{j}$ is set to $100.$
The fitness function $f$ is defined as:

\[
f=\frac{1}{1+10\sum error_{j}}. 
\]

If $f$ is $1,$ that is, if the circuit is correct, the bonus of $1/\left(
number-of-gates\right) $ is added to $f$ so as to apply a selection pressure
based upon the circuit size (fitness is enlarged).

In \cite{yabuki00genetic} authors used the roulette-wheel selection and
two-point crossover with probability $0.7$. Differently from the work found
in \cite{rubinstein00evolving}, for circuit design (section \ref
{Quantum-Design}), mutation is considered more significant in this case. The
mutation probability is $1/\left( chromosome-length\right) $; that is, the
algorithm is biased in the preservation of smaller chromosomes against the
larger ones. The population size and the maximum number of generations was $%
5,000$ and $1,000$, respectively. All individuals are replaced every
generation (there is no elitism). The simpler circuit so obtained is
pictured on Figure \ref{fig3-design-qc}:

\begin{figure}[tbph]
\epsfxsize=8.0cm
\par
\begin{center}
{\mbox{\epsffile{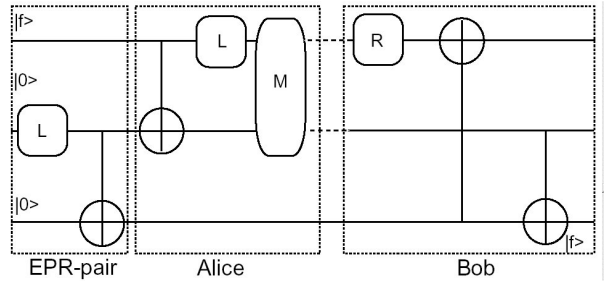}}}
\end{center}
\caption{Optimized circuit for the quantum teleportation problem.}
\label{fig3-design-qc}
\end{figure}

We can check that the circuit on Figure \ref{fig3-design-qc} is encoded by
the string given by (\ref{string000}). This circuit has $8$ gates while
Brassard's had eleven (Figure \ref{fig2-design-qc}), which demonstrate the
capabilities of the GA procedure to evolve an initial circuit towards a
simpler one.

We are in charge with an implementation of this algorithm, but using
different strategies. We shall return to this point in section \ref
{Discussion}.

\section{Quantum Evolutionary Computation \label{QEC}}

\subsection{Quantum-Inspired Genetic Algorithms \label{QIGAs}}

In this section we present the second part of our review: the analysis of
genetic algorithms based on quantum computing concepts. This is an important
step towards the implementation of genetic algorithms in a quantum hardware.

We start with the Quantum-Inspired Genetic Algorithm (QIGA) proposed in \cite
{han00genetic}. The QIGA is characterized by principles of quantum computing
including qubits and probability amplitude. It uses a qubit representation
instead of the usual binary, numeric, or symbolic representations \cite
{Koza1992,Mitchell1996}. More specifically, QIGA uses a \textit{m-qubit}
representation, defined as:

\begin{equation}
\left[ \left( 
\begin{array}{l}
\alpha _{10} \\ 
\alpha _{11}
\end{array}
\right) ,\left( 
\begin{array}{l}
\alpha _{20} \\ 
\alpha _{21}
\end{array}
\right) ,...,\left( 
\begin{array}{l}
\alpha _{m0} \\ 
\alpha _{m1}
\end{array}
\right) \right] ,  \label{qiga00}
\end{equation}
where each pair $\left( \alpha _{i0},\beta _{i1}\right); i=1,...,m $,
indicates a qubit.

Now, we must explain how convergence can be obtained with the qubit
representation. Let us consider the following scheme, which is proposed in 
\cite{han00genetic}.

For each \textit{m}-qubit chromosome of the form (\ref{qiga00}), a binary
string $\left\{ x_{1},x_{2},...,x_{m}\right\} $ is defined, where each bit
is selected using the corresponding qubit probability, $\left| \alpha
_{i0}\right| ^{2}$ or $\left| \alpha _{i1}\right| ^{2}$. Observe that if $%
\left| \alpha _{i0}\right| ^{2}$ or $\left| \alpha _{i1}\right| ^{2}$
approaches to $1$ or $0$, the qubit chromosome converges to a single state
and the diversity given by the superposition of states disappears gradually.

An application dependent fitness function is used to evaluate the solution $%
\left\{ x_{1},x_{2},...,x_{m}\right\} $. Another step is to design efficient
evolutionary strategies. This would be accomplished through crossover and
mutations but their implementations are not explained in \cite{han00genetic}
. Obviously, as usual, we can suppose a one-point crossover between parent
chromosomes as well as unitary operators to change a randomly chosen qubit $%
\left( \alpha _{i0},\beta _{i1}\right) $ of the expression (\ref{qiga00}).
However, as the QIGA has diversity caused by the qubit representation, the
role of genetic operators is not clear. Also, it is stated in \cite
{han00genetic} that, if the probabilities of mutation and crossover are
high, the performance of the QIGA can be decreased notably.

At the beginning of the algorithm, a population $Q\left( t\right) =\left\{
q_{1}^{t},q_{2}^{t},...,q_{m}^{t}\right\} $ of \textit{m}-qubit chromosomes
is instantiated. Given a \textit{m}-qubit chromosome in $Q\left( t\right) $
we can find the corresponding binary string through the rule stated above.
The so obtained binary string population will be denoted by $P\left(
t\right) .$

Besides, there is an update step which aims to increase the probability of
some states. Henceforth, given a qubit $\left( \alpha _{i0},\alpha
_{i1}\right) $ of a \textit{m}-qubit chromosome, it is updated by using the
rotation gate $U\left( \theta _{i}\right) $:

\begin{equation}
U\left( \theta _{i}\right) =\left[ 
\begin{array}{cc}
\cos \left( \theta _{i}\right)  & -\sin \left( \theta _{i}\right)  \\ 
\sin \left( \theta _{i}\right)  & \cos \left( \theta _{i}\right) 
\end{array}
\right] ;\quad \left[ 
\begin{array}{c}
\alpha _{i0}^{\prime } \\ 
\alpha _{i1}^{\prime }
\end{array}
\right] =U\left( \theta _{i}\right) \left[ 
\begin{array}{c}
\alpha _{i0} \\ 
\alpha _{i1}
\end{array}
\right] ,  \label{rotate00}
\end{equation}
where $\theta _{i}$ is formed through the binary solutions $P\left( t\right) 
$ and the best solution found (see next).

Let us present a pseudo-code of the QIGA developed in \cite{han00genetic}:

\textbf{Procedure QIGA}

$begin$

\qquad $t\leftarrow 0$

$\qquad $Initialize $Q\left( t\right) $

$\qquad $Make $P\left( t\right) $ by observing $Q\left( t\right) $

$\qquad $Evaluate $P\left( t\right) $

$\qquad $Store the best solution $b$ among $P\left( t\right) $

\qquad $while$(not termination-condition) do

\qquad $begin$

$\qquad \qquad t\leftarrow t+1$

$\qquad $\qquad Make $P\left( t\right) $ by observing $Q\left( t-1\right) $

$\qquad $\qquad Evaluate $P\left( t\right) $

$\qquad $\qquad Update $Q\left( t\right) $ using quantum gates $U\left(
t\right) $

$\qquad $\qquad Store the best solution $b$ among $P\left( t\right) $

\qquad $end$

$end$

The quantum gates $U\left( t\right) $ are application dependent. This step
aims to improve the convergence. After updating $Q\left( t\right) ,$ the
best solution among $P\left( t\right) $ is selected, and if the solution is
fitter than the stored best solution, the stored solution is replaced by the
new one. The binary solutions $P\left( t\right) $ are discarded at the end
of the loop. A parallel version of the QIGA is presented in \cite
{Parallel-han00genetic}.

A significant point to be considered is the exploration of the tensor
product to enlarge diversity. Despite of authors claim in \cite{han00genetic}%
, the scheme proposed did not take advantage of such effects at all. We will
analyze this point in section \ref{Discussion}.

\subsubsection{Experiments for QIGA \label{QIGA-Exper}}

The knapsack problem, which is a kind of combinatorial optimization problem 
\cite{Nemhauser1998}, is used in \cite{han00genetic} to investigate the
performance of QIGA. The $0-1$ knapsack problem is described as: given a set
of $m$ items and a knapsack with limited capacity $C$, select a subset of
the items so as to maximize a profit function $f\left( x\right) $ given by:

\begin{equation}
f\left( x\right) =\sum_{i=1}^{m}p_{i}x_{i},  \label{qiga003}
\end{equation}
and subjected to:

\begin{equation}
\sum_{i=1}^{m}w_{i}x_{i}<C,  \label{qiga004}
\end{equation}
where $\left( x_{1},x_{2},...,x_{m}\right) \in \left\{ 0,1\right\} ^{m}$, $%
p_{i}$ and $w_{i}$ are the profit and the weight associated to the item $i$,
respectively.

When applying the QIGA to this problem, the length of a qubit chromosome is
the same as the number of items. The $i-th$ item can be selected for the
knapsack with probability $\left| \alpha _{i0}\right| ^{2}$, following the
procedure given in section \ref{QIGAs}. Thus, from each \textit{m}-qubit
chromosome a binary string of the length $m$ is formed. The binary string $%
x_{j}$ represents the $j-th$ candidate solution to the problem. The $i-th$
item is selected for the knapsack if and only if $x_{ij}=1.$ To measure the
efficiency of the QIGA, its performance was compared with that one of
conventional genetic algorithms (CGAs). Three types of CGAs were considered:
algorithms based on penalty functions, algorithms based on repair methods,
and algorithms based on decoders \cite{han00genetic,Nemhauser1998}.

For the first group of algorithms, the profit function is:

\begin{equation}
f\left( x\right) =\sum_{i=1}^{m}p_{i}x_{i}-Pen\left( x\right) ,
\label{qiga008}
\end{equation}
where $Pen\left( x\right) $ is a penalty function. Among the possibilities
to define penalty functions the following ones were considered in \cite
{han00genetic}:

\begin{equation}
Pen_{1}\left( x\right) =\log _{2}\left( 1+\rho \left(
\sum_{i=1}^{m}w_{i}x_{i}-C\right) \right) ,  \label{qiga009}
\end{equation}

\begin{equation}
Pen_{2}\left( x\right) =\rho \left( \sum_{i=1}^{m}w_{i}x_{i}-C\right) ,
\label{qiga0010}
\end{equation}

\begin{equation}
Pen_{3}\left( x\right) =\left( \rho \left( \sum_{i=1}^{m}w_{i}x_{i}-C\right)
\right) ^{2}.  \label{qiga0011}
\end{equation}
where $\rho =\max \left\{ p_{i}/w_{i};\quad i=1,...,m\right\} $.

For repair methods, the profit is defined by:

\begin{equation}
f\left( x\right) =\sum_{i=1}^{m}p_{i}x_{i}^{\prime },  \label{qiga0013}
\end{equation}
where $x^{\prime }$ is a repaired vector of the original vector $x$.
Original chromosomes are replaced with a $5\%$ probability in the
experiment. The two repair algorithms considered in \cite{han00genetic}
differ only in selection procedure, which chooses an item for removal from
the knapsack:

$Rep_{1}$(random repair): The selection procedure selects a random element
from the knapsack.

$Rep_{2}$(greedy repair): All items in the knapsack are sorted in the
decreasing order of their profit to weight ratios. The selection procedure
always chooses the last item for deletion.

A possible decoder for the knapsack problem is based on an integer
representation. Each chromosome is a vector of $m$ integers; the $i-th$
component of the vector is an integer in the range from $1$ to $m-i+1$. The
ordinal representation references a list $L$ of items; a vector is decoded
by selecting appropriate item from the current list. The two algorithms for
this class used in \cite{han00genetic} are:

$Dec_{1}$(random decoding): The build procedure creates a list $L$ of items
such that the order of items on the list corresponds to the order of items
in the input file which is random.

$Dec_{2}$(greedy decoding): The build procedure creates a list $L$ of items
in the decreasing order of their profit to weight ratios.

Besides, there were an experiment that implements a scheme using $Pen_{2}$
and $Rep_{1}$.

The QIGA proposed in \cite{han00genetic} for this problem contains a repair
algorithm. It can be described as follows:

\textbf{Procedure QIGA-Knapsack}

$begin$

\qquad $t\leftarrow 0$

$\qquad $Initialize $Q\left( t\right) $

$\qquad $Make $P\left( t\right) $ by observing $Q\left( t\right) $

$\qquad $repair $P\left( t\right) $

$\qquad $Evaluate $P\left( t\right) $

$\qquad $Store the best solution $b$ among $P\left( t\right) $

\qquad $while$($t<MAX\_GEN$) do

\qquad $begin$

$\qquad \qquad t\leftarrow t+1$

$\qquad $\qquad Make $P\left( t\right) $ by observing $Q\left( t-1\right) $

$\qquad $\qquad repair $P\left( t\right) $

$\qquad $\qquad Evaluate $P\left( t\right) $

$\qquad $\qquad Update $Q\left( t\right) $ using quantum gates $U\left(
t\right) $

$\qquad $\qquad Store the best solution $b$ among $P\left( t\right) $

\qquad $end$

$end$

\textbf{Procedure make}$\left( x\right) $

$begin$

\qquad $i\leftarrow 0$

\qquad $while$($i<m$) do

\qquad $begin$

\qquad \qquad $i\leftarrow i+1$

\qquad \qquad if $random\left[ 0,1\right] >\left| \alpha _{i0}\right| ^{2}$

\qquad \qquad \qquad then $x_{i}\leftarrow 1$

\qquad \qquad else $x_{i}\leftarrow 0$

\qquad $end$

$end$

\textbf{Procedure repair}$\left( x\right) $

$begin$

\qquad knapsack-overfilled $\leftarrow $ false

\qquad if $\sum_{i=1}^{m}w_{i}x_{i}>C$

\qquad then knapsack-overfilled $\leftarrow $ true

\qquad while (knapsack-overfilled) do

\qquad $begin$

\qquad \qquad select an $i-th$ item from the knapsack

\qquad \qquad $x_{i}\leftarrow 0$

\qquad \qquad if $\sum_{i=1}^{m}w_{i}x_{i}\leq C$

\qquad \qquad then knapsack-overfilled $\leftarrow $ false

\qquad $end$

\qquad while (not knapsack-overfilled) do

\qquad $begin$

\qquad \qquad select a $j-th$ item from the knapsack

\qquad \qquad $x_{j}\leftarrow 1$

\qquad \qquad if $\sum_{i=1}^{m}w_{i}x_{i}>C$

\qquad \qquad then knapsack-overfilled $\leftarrow $ true

\qquad $end$

\qquad $x_{j}\leftarrow 0$

$end$

The profit of a binary solution is evaluated by expression (\ref{qiga0013})
and it is used to find the best solution $b$ among $P\left( t\right) $. A 
\textit{m}-qubit chromosome is updated by using the rotation gates,
following expression (\ref{rotate00}). The angles $\theta _{i}$ are computed
as follows. Let us suppose that we have a binary string $x=\left(
x_{1},x_{2},...,x_{m}\right) $ such that $f\left( x\right) >f\left( b\right) 
$, where $f$ is defined by expression (\ref{qiga0013}). If $x_{i}=1$ and $%
b_{i}=0$, the idea is to set the value of $\theta _{i}=sign\left( \alpha
_{i0}\cdot \alpha _{i1}\right) \Delta \theta _{i}$ such that the probability
amplitude of $\mid 1\rangle $ is increased. Thus, we want that $\left|
\alpha _{i1}^{\prime }\right| ^{2}>\left| \alpha _{i1}\right| ^{2}$, where $%
\alpha _{i1}^{\prime }$ is given by equation (\ref{rotate00}). So:

\begin{equation}
\left| \alpha _{i1}^{\prime }\right| ^{2}=\left( \alpha _{i0}\sin \theta
_{i}+\alpha _{i1}\cos \theta _{i}\right) ^{2}=\left( \alpha _{i0}\sin \theta
_{i}\right) ^{2}+\left( \alpha _{i1}\cos \theta _{i}\right) ^{2}+2\alpha
_{i0}\alpha _{i1}\sin \theta _{i}\cos \theta _{i},  \label{eq000}
\end{equation}
where we have supposed that $\alpha _{i0},\alpha _{i1}$ are real ones for
simplicity. Thus, to increase the desired probability amplitude as much as
possible we should set $sign\left( \alpha _{i0}\cdot \alpha _{i1}\right)
=+1,-1,0,$ according to $\alpha _{i0}\alpha _{i1}>0,$ $\alpha _{i0}\alpha
_{i1}<0,$ or $\alpha _{i0}\alpha _{i1}=0$, respectively. The setting of $%
\Delta \theta _{i}$ is through experimentation. In the reported example, it
was set to $0.025$. Following such procedure, a lookup table for $\theta
_{i} $ can be performed (see \cite{han00genetic} details).

The update procedure is given bellow:

\textbf{Procedure update}$\left( q\right) $

$begin$

\qquad $i\leftarrow 0$

\qquad while ($i<m$) do

\qquad $begin$

\qquad \qquad $i\leftarrow i+1$

\qquad \qquad determine $\theta _{i}$

\qquad \qquad obtain $q^{\prime }=\left( \alpha _{i0}^{\prime },\alpha
_{i1}^{\prime }\right) $ as:

\qquad \qquad $\left( \alpha _{i0}^{\prime },\alpha _{i1}^{\prime }\right)
^{T}=U\left( \theta _{i}\right) \left( \alpha _{i0},\alpha _{i1}\right) ^{T}$

\qquad $end$

\qquad \qquad $q\leftarrow q^{\prime }$

$end$

The results obtained by the QIGA\ just presented, reported in \cite
{han00genetic}, uses the following profits and weights:

\begin{eqnarray}
w_{i} &=&uniformly-random[1,10),  \label{qiga006} \\
pi &=&w_{i}+5.  \nonumber
\end{eqnarray}

The average knapsack capacity was used:

\begin{equation}
C=\frac{1}{2}\sum_{i=1}^{m}w_{i}.  \label{qiga007}
\end{equation}

The data files were unsorted and the number of items were $100,$ $250$ and $%
500$.

The population size of the eight conventional genetic algorithms was equal
to $100$. Probabilities of crossover and mutation were fixed: $0.65$ and $%
0.05$, respectively. The population size is $1$, for the first series of
experiments, and $10$ for the second one. As a performance measure of the
algorithm the best solution found within $500$ generations over $25$ runs is
collected. Also, the elapsed time per one run is checked.

For $100$ items QIGAs yielded superior results as compared to all the CGAs.
For $250$ and $500$ items the QIGA with $10$-size population outperforms all
the classical ones \cite{han00genetic}.

\subsection{Quantum Genetic Algorithms \label{QGAs}}

The work reported on section \ref{QIGAs} shows that the application of
quantum computing concepts to evolutionary programming is a promising
research. The results presented points out that a quantum genetic algorithm
(QGA) would outperform the classical ones. Besides, such implementation
would take advantage of quantum parallelism as well as GAs parallelism. The
obvious question is how to implement genetic algorithms in quantum computers?

The reference \cite{rylander:2001:gecco} is an effort to produce a QGA.
Despite of the fact that there are several open points, it is the first
effort in the direction of such algorithm.

\begin{sloppypar}The QGA proposed in \cite{rylander:2001:gecco} uses two registers for each 
\textit{quantum individual}; the first one stores an \textit{\ individual }%
while the second one stores the individual's fitness. These two registers
are referred as \textit{individual register} and the \textit{fitness register}%
, respectively. A population of $N$ quantum individuals is stored through
pairs of registers $\left(
individual-register_{i},fitness-register_{i}\right) $, $i=1,..,N$.\end{sloppypar} 

At different times during the QGA the fitness register would store a single
fitness value or a quantum superposition of fitness values. Identically for
the individual register.

Once a new population is generated, the fitness for each individual would be
calculated and the result stored in the individual 's fitness register.

The effect of the fitness measurement is a collapse given by expression (\ref
{measure001}). This process reduces each quantum individual to a
superposition of classical individuals with a common fitness. It is a key
step in the QGA \cite{rylander:2001:gecco}. Then, crossover and mutation
would be applied. The whole algorithm can be written as follows:

\textbf{Quantum Genetic Algorithm}

\textbf{Generate} a population of quantum individuals.

\textbf{Calculate} the fitness of the individuals.

\textbf{Measure} the fitness of each individual (collapse).

$while$(termination-condition) do

\qquad \textbf{Selection} based on the observed fitness.

\qquad \textbf{Crossover} and \textbf{Mutations } are applied.

\qquad \textbf{Calculate} the fitness of the individuals.

\qquad \textbf{Measure} the fitness of each individual (collapse).

$end\quad while$

According to \cite{rylander:2001:gecco}, the more significant advantage of
QGA's will be an increase in the production of good building blocks (\textit{%
schemata }\cite{Holland1975,Mitchell1996}) because, during the crossover,
the building block is crossed with a superposition of many individuals
instead of with only one in the classical GAs.

One can also view the evolutionary process as a dynamic map in which
populations tend to converge on fixed points in the population space. From
this viewpoint the advantage of QGA is that the large effective size allows
the population to sample from more basins of attraction. Thus, it is much
more likely that the population will include members in the basins of
attraction for the higher fitness solutions.

Another advantage is the quantum computer's ability to generate true random
numbers. By applicating Kolmogorov complexity analysis, it has been shown
that the output of classical implementations in genetic programming, which
use a pseudo random number generator, are bounded above by the genetic
programming itself, whereas with the benefit of a true random number
generator there is no such bound \cite
{rylander:2001:gecco,Rylander:2001:CCGPaI}.

Despite of these promising features, fundamental points are not addressed in 
\cite{rylander:2001:gecco}. Firstly, it is not clear how to implement
crossover in a quantum computers. Besides, how to perform the fitness
function calculation in quantum hardware? Even a much more fundamental
problem is that to explore the superposition of quantum individuals the
correlation $individual\leftrightarrow fitness$ must be kept during the
whole computation. Entanglement seems to be the only possibility to
accomplish this task. But, in this case, things must be formally described
to avoids misunderstandings and wrong interpretations. We develop such
mathematical formalism on section \ref{Discussion}.

\section{Discussion and Perspectives \label{Discussion}}

In this section we analyze some issues concerning to the reviewed methods.
Possible solutions and perspectives in this area are also discussed.

In \cite{Thess2003} we show some challenges concerning the GA for learning
linear operators (section \ref{OurGA}). Other tests presented in \cite
{Thess2003} show that the number of generations seems to increase when space
dimension gets higher. The increasing rate must be controlled if we change
the population size properly. However, such procedure could be a serious
limitation of the algorithm for large linear systems.

\begin{sloppypar}The behavior for underconstrained problems; that is, when $S=\left\{ \left(
\mid \chi _{i}\rangle ,\mid \psi _{i}\rangle \right) ;\quad F\mid \chi
_{i}\rangle =\mid \psi _{i}\rangle ,\quad i=1,...,K<n\right\} ,$ where $n$
is the space dimension, is also analyzed in \cite{Thess2003}. In this case,
we had to increase the population size but the number of generations is
smaller than that one for the constrained test ($K=n$). As we expect, there
is a trade-off between the increase of solutions and the fact that we are
less able to properly evolve the populations due to the lack of prior
information. Moreover, the observed success is an advantage of the
method, if compared with traditional ones. In this case, numerical
approaches based on iterative methods in matrix theory (Gauss-Seidel, GMRES,
etc) can not be applied without extra machinery because the solution is not
unique \cite{Thess2003}. \end{sloppypar}

The comparison with Dan Ventura's learning method, given on section \ref
{Experimental}, shows that our algorithm overcomes the limitation of the
later: we do not need that $\left\{ \mid \chi _{0}\rangle ,\mid
\chi_{1}\rangle \right\}$ is an orthonormal basis of the vector space.
However, when using our GA method, we pay a price due to storage
requirements and computational complexity.

Dan Ventura's algorithm as well as numerical methods (see \cite{Thess2003}
and references therein), have a computational cost asymptotically limited by 
$O\left( n*n^{2}\right) $ while our GA method needs $O\left(
Ngen*N*n^{2}\right) $ float point operations. Besides, for traditional
numerical methods and Ventura's algorithm, we observe a storage requirements
of $O\left( n^{2}\right) $ against $O\left( N*n^{2}\right) $ for our
approach. Thus, the disadvantage of our method becomes clear.

However, if compared with matrix methods, our algorithm is in general less
sensitive to roudoff errors \cite{Golub1985}. This is due to, unlikely
numerical methods that try to follow a path linking the initial position to
the optimum, our GA algorithm searches the solution through a set of
candidates.

\begin{sloppypar}To improve the convergence we need better evolutionary
(crossover/mutation) strategies. The behavior pictured on Figure \ref{caso2x2Ventura}, 
of section \ref{Experimental}, is a typical one for every test we made \cite{Thess2003}. 
It indicates that our evolutionary strategies are
efficient to get closer the solution but not to complete the learning
process. Further analysis should be made to improve these 
operators.\end{sloppypar}

When comparing the works for quantum circuit design, \cite
{rubinstein00evolving} and \cite{yabuki00genetic}, we observe the following
aspects:

1) Gate representation: The gate structure of \cite{rubinstein00evolving}
versus the codon used in \cite{yabuki00genetic}. Despite of some apparent
difference between them, it is simple to check that they are equivalent, in
the sense that, any gate can be represented with either the later or the
former.

2) Genetic Operators: Both implementations have used crossover and
mutations. However, the crossover implementation used in \cite
{rubinstein00evolving} operates on all the levels of an individuals
structure (the gates, each category of qubit operands and each parameter
type) while in \cite{yabuki00genetic} it affects only the gate and qubit
levels. Mutations are basically equivalent because, if the alleles are
randomly changed, like in \cite{yabuki00genetic}, we are randomly replacing
a gate with a new one, like in \cite{rubinstein00evolving}, and vice-versa.

3) Range of Applications: Despite of the fact that the aim of \cite
{yabuki00genetic} is circuit optimization, it can be straightforwardly
adapted for circuit design. This can be accomplished by changing the
mutation probability (the formula $1/\left( chromosome-length\right) $ does
not make sense in this case). We can follow \cite{rubinstein00evolving} and
set this probability to a small value (typically $0.001$). Besides, the
fitness function remains case dependent and we stop using the bonus $%
1/\left( number-of-gates\right) $ to bias the solution to smaller circuits
(if we do not know any prior correct circuit, there is no sense for
prefering smaller circuits over bigger ones during evolution). Besides, some
elitism may be introduced. Now, we are analyzing such modifications.

Moreover, a more fundamental question about GAs for circuit design follows
from the next comments. The algorithms \cite
{rubinstein00evolving,yabuki00genetic} basically evolve an initial
population of individuals towards a desired circuit. Evolution can be
regarded as the exploration of search spaces by populations. Thus, an
interesting question in this case is what about the structure of such spaces
for circuit designing/optimization?

For instance, we must observe that without the identity operator the circuit
size (number of gates) will be a variable. Henceforth, a search space of $12$%
-codons strings (like expression (\ref{string000})), would be transformed in
another one with just $8$-codons strings at the end of the optimization
process. This can be seen as an evolutionary process called \textit{%
innovation} \cite{Wagner2000}.

Following \cite{Wagner2000} we do need a mathematical representation in
which the kind and number of codons follow from the dynamics of the model.
In \cite{Wagner2000} the concept of configuration spaces is proposed as one
of such approach. Obviously, the Identity operator is a simple trick to
address this problem if we know in advance the maximum circuit size.
However, the concept of configuration spaces might open possibilities to
analyze the structure of the circuit space. That is way we are going to
consider this mathematical framework in our research.

A configuration space is a set of objects (circuits, for example) as well as
a topological structure on this set which describes how these objects can be
transformed into each other by an operator \cite{Wagner2000}.

Symmetries of the configuration space induced by evolutionary mechanisms
(mutation and crossover, for instance) are fundamental elements in this
framework. They define the dimensionality of the space in which evolution
occurs. Hence, any evolutionary process that affects the symmetries of the
configuration space may change its dimensionality (the number of
non-identity gates, in our case). The configuration space formalism includes
beautiful mathematical results in finite Abelian groups \cite{Wagner2000}.
We wish to analyze the algorithms proposed in \cite
{rubinstein00evolving,yabuki00genetic} through this formalism, in order to
derive more efficient evolutionary strategies.

When considering the QIGA presented on section \ref{QIGAs}, some
explanations must be offered about the diversity that can be achieved by the 
\textit{m-}qubit representation \cite{han00genetic}.

Thus, let us take the \textbf{make}$\left( x\right) $ procedure. For
simplicity, we are going to consider a $3$-qubit chromosome. Let $%
r_{1},r_{2},r_{3}$ be the random numbers generated during the $while$ loop
execution in \textbf{make}$\left( x\right) $. We could have $r_{1}>\left|
\alpha _{10}\right| ^{2},r_{2}>\left| \alpha _{20}\right| ^{2},r_{3}<\left|
\alpha _{30}\right| ^{2}$. Thus, the generated string would be $\left(
1,1,0\right) $.

Now, consider the tensor product

\begin{equation}
\mid \psi _{1}\rangle \otimes \mid \psi _{2}\rangle \otimes \mid \psi
_{3}\rangle =\sum_{i_{1},i_{2},i_{3}\in \left\{ 0,1\right\} }^{{}}\alpha
_{1i_{1}}\alpha _{2i_{2}}\alpha _{3i_{3}}\mid i_{1}\rangle \otimes \mid
i_{2}\rangle \otimes \mid i_{3}\rangle .  \label{qiga001}
\end{equation}

Thus, the qubit chromosome will be represented as a superposition of the
states $\mid i_{1}\rangle \otimes \mid i_{2}\rangle \otimes \mid
i_{3}\rangle ,\quad i_{1},i_{2},i_{3}\in \left\{ 0,1\right\} $, and so it
carries information about all of them at the same time. Such observation
points out the fact that the qubit representation has a better
characteristic of diversity than classical approaches, since it can
represent superposition of states. In classical representations we will need
at least $2^{3}$ chromosomes to keep the information carried by expression (%
\ref{qiga001}) while only one 3-qubit chromosome is enough.

However, the probability amplitude of the state $\mid 1\rangle \otimes \mid
1\rangle \otimes \mid 0\rangle $ may not be the largest one. Henceforth, it
does not seems that the binary string generation rule proposed in \cite
{han00genetic} does explore such diversity in general.

However, if we take another generation rule, say: if $\left| \alpha
_{i1}\right| ^{2}>\left| \alpha _{i0}\right| ^{2}$ then $x_{i}\leftarrow 1$,
else $x_{i}\leftarrow 0$, thus we can be sure that the generated binary
string is an index to the larger amplitude probability of state (\ref
{qiga001}). Experiments must be performed in order to show the efficiency of
such rule.

The QGA presented in \cite{rylander:2001:gecco} exploits the quantum effects
of superposition and entanglement. However, the lack of a more formal
explanation raises some questions. How to implement crossover in quantum
computers? How to compute the fitness function? What about a mathematical
definition of a quantum individual? These are examples of such questions. 

Now, we address some of these points in order to be closer to answer the
question: what GAs will look like as an implementation on quantum hardware?

The starting point of our development comes from the known problem of
finding the period $r$ of a periodic function $f:Z_{N}\rightarrow Z$, where $%
Z_{N}$ denotes the additive group of integers modulo $N$.

In this case, the quantum solution provided by Shor \cite{Chuang2000} uses a
hardware with two registers in the following entangled state:

\begin{equation}
\mid \Psi \rangle =\frac{1}{\sqrt{N}}\sum_{x=0}^{N-1}\mid x\rangle \otimes
\mid f\left( x\right) \rangle .  \label{entan01}
\end{equation}

Thus, according to the expression (\ref{measure001}), by measuring the
second register, yielding, say, a value $y_{0}$, the first register 's state
will collapse to an uniform superposition of all those $\mid x\rangle
^{\prime }s$ such that $f\left( x\right) =y_{0}$; that is:

\begin{equation}
\mid \Psi \rangle _{after}=\frac{1}{\sqrt{K}}\sum_{k=0}^{K-1}\mid
x_{0}+kr\rangle ,  \label{period00}
\end{equation}
where $x_{0}$ is such a $x$ and $N=Kr$. When using the state $\mid \Psi
\rangle ,$ given by expression (\ref{entan01}),  the desired effect is to
keep the correspondence between each integer $x$ with its corresponding
value $f\left( x\right) $.

Now, let us return to the QGA of section \ref{QGAs} and present a physical
description of it. The quantum individual could be mathematically
represented by a state given by expression (\ref{entan01}), where $\mid
x\rangle $ represents an individual and $f\left( x\right) $ its fitness.
Thus, we keep the idea of representing a quantum individual through two
registers which was used in \cite{rylander:2001:gecco}. So, if we have $M$
quantum individuals in each generation we need $M$ register pairs (\textit{%
individual register}, \textit{fitness register}).

In our formulation, each register is a closed quantum system. Thus all of
them can be initialized with the state given by expression (\ref{entan01}).
Then, unitary operators $W$ will be applied in order to complete the
generation of the initial population. Henceforth, the initialization could
encompass the following steps:

1) For each register $i,$ generate the state:

\[
\mid \varphi \rangle _{i}=\frac{1}{\sqrt{N}}\sum_{x=0}^{N-1}\mid x\rangle
_{i}\otimes \mid 0\rangle _{i},\quad i=1,...,M, 
\]

2) Apply unitary operators $W$ (rotations, for example) and $U_{f}$, the
known black box which performs the operation $U_{f}\mid a\rangle \otimes
\mid 0\rangle =\mid a\rangle \otimes \mid f\left( a\right) \rangle $ \cite
{Chuang2000}, to complete the initial population:

\[
\mid \Psi \rangle _{i}\equiv U_{f}W\mid \varphi \rangle
_{i}=\sum_{x=0}^{N-1}U_{f}\left( W\left( \frac{\mid x\rangle _{i}}{\sqrt{N}}%
\right) \otimes \mid 0\rangle _{i}\right) = 
\]

\[
=\sum_{x=0}^{N-1}U_{f}\left( a_{x}\mid x\rangle _{i}\otimes \mid 0\rangle
_{i}\right) = 
\]

\begin{equation}
=\sum_{x=0}^{N-1}a_{xi}\mid x\rangle _{i}\otimes \mid f\left( x\right)
\rangle _{i},\quad i=1,...,M.  \label{individual000}
\end{equation}

We must highlight that all the above operations are unitary ones,
consequently, can be performed in quantum computers \cite{Oskin2002}.
Besides, it is important to observe that the fitness is stored in the second
register after the generation of the population. Now, by measuring the
fitness, each individual undergoes collapse, according to the expression (%
\ref{period00}):

\begin{equation}
\mid \Psi \rangle _{i}^{after}=\frac{1}{\sqrt{K_{i}}}\sum_{k=0}^{K_{i}-1}%
\mid k\rangle _{i}\otimes \mid y_{0}\rangle _{i},  \label{qga002}
\end{equation}
where $\mid k\rangle _{i}$ is such that the observed fitness for the $i-th$
register is $f\left( k\right) =y_{0}$.

When entering the main loop, the observed fitness is used to select the best
individuals. Then, genetic operators must be applied.

Mutations can be implemented through the following steps.

1) Apply $U_{f}^{-1}$ over the measurement result:

\begin{equation}
U_{f}^{-1}\mid \Psi \rangle _{i}^{after}=\frac{1}{\sqrt{K_{i}}}%
\sum_{k=0}^{K_{i}-1}\mid k\rangle _{i}\otimes \mid 0\rangle _{i},
\label{mutations000}
\end{equation}

2) Unitary operators $P$ (small rotations, for example) are applied to the
above result:

\[
P\left( U_{f}^{-1}\mid \Psi \rangle _{i}^{after}\right)
=\sum_{k=0}^{K_{i}-1}P\left( \frac{\mid k\rangle _{i}}{\sqrt{K_{i}}}\right)
\otimes \mid 0\rangle _{i}= 
\]

\begin{equation}
=\sum_{x=0}^{N-1}\beta _{xi}\mid x\rangle _{i}\otimes \mid 0\rangle _{i},
\label{mutation0001}
\end{equation}
where we expanded the result in the computational basis.

3) Finally, apply $U_{f}$ to recover the diversity that was lost during the
measurement:

\begin{equation}
U_{f}PU_{f}^{-1}\mid \Psi \rangle _{i}^{after}=\sum_{x=0}^{N-1}\beta
_{xi}\mid x\rangle _{i}\otimes \mid f\left( x\right) \rangle _{i}.
\label{mutation0002}
\end{equation}

The development given above allows to discuss some points. Firstly, we
observe that if we take a superposition of individuals in the first register
and the corresponding fitness superposition in the second one, as claimed in 
\cite{rylander:2001:gecco}, we will have:

\[
\mid \Psi \rangle =\left( \sum_{x=0}^{N-1}a_{x}\mid x\rangle \right) \otimes
\left( \sum_{x=0}^{N-1}b_{x}\mid f\left( x\right) \rangle \right)
=\sum_{x=0}^{N-1}\sum_{y=0}^{N-1}a_{x}b_{y}\mid x\rangle \otimes \mid
f\left( y\right) \rangle . 
\]

Thus, we are not able to keep the correlation $individual\leftrightarrow
fitness.$ For instance, after a measurement that gives a $z_{0}$ value, the
system state would be:

\[
\mid \Psi \rangle _{{}}^{after}=\sum_{x=0}^{N-1}\mid x\rangle \otimes \mid
f\left( y_{0}\right) \rangle , 
\]
where $f\left( y_{0}\right) =z_{0}$ (observe that in general $x\neq y_{0}$
in this expression). So, such proposal does not seems to be efficient at all.

According to \cite{rylander:2001:gecco}, the major advantage for a QGA is
the increased diversity of a quantum population due to superposition, which
we have precisely defined through expression (\ref{individual000}). This
effective size decreases during the measurement of the fitness, when the
superposition is reduced to only individuals with the observed fitness,
according to expression (\ref{qga002}). However, it would be increased
during the crossover and mutation applications. Besides, by increasing
diversity it is much more likely that the population will include members in
the basins of attraction for the higher fitness solutions. Thus, an improved
convergence rate must be expected. Besides, classical individuals with high
fitness can be relatively incompatible; that is that any crossover between
them is unlikely to produce a very fit offspring. However, in the QGA, these
individuals can co-exist in a superposition.

Despite of the mathematical development given above, two fundamental points
remain. Firstly, we can not suppose that the number of elements of the
search space is the same of the number of states of the computational basis (%
$N$, in the above presentation). If so, the solution would be to find the
maximum value of $f\left( x\right) $, which is just an optimization problem
that can be addressed by quantum optimization algorithms \cite
{Protopopescu2003}. Besides, the search space size is in general too large
that makes some assumption unreasonable.

Secondly, the crossover needs special considerations not only because
combination of states in Hilbert spaces is limited by the constraint of
unitary operations but also because each register pair is a closed quantum
system. Thus, we need some kind of \textit{quantum communication channel} to
combine states. This question should be addressed in the context of
state-of-the-art quantum computers architecture (see \cite{Oskin2002} and
references therein).

\section{Conclusions \label{Concl}}

In this paper we survey the main works in quantum evolutionary programming
and in the applications of GAs to address some problems in quantum
computation. Besides, we offer new perspectives in the area which are part
of our current research in this field. Among them, we believe that the
analysis of the algorithms proposed in \cite
{rubinstein00evolving,yabuki00genetic} through the configuration space
formalism and a QGA\ implementation are the most exciting ones.

The concept of configuration spaces might open possibilities to analyze the
structure of the circuit space in order to derive more efficient
evolutionary strategies.

On the other hand, a QGA implementation could take advantage of both the
quantum computing and GAs parallelism. We analyze the work summarized on
section \ref{QGAs} and give a formal explanation of its main elements.
However, quantum crossover and efficient strategies for search space
exploration remains challenges in this field.

\section{Acknowledgments}

We would like to acknowledge PIBIC-LNCC for the financial support for this
work.

\end{document}